\documentclass[10pt,journal,cspaper,compsoc]{IEEEtran}
\ifCLASSOPTIONcompsoc
\else
\fi
\ifCLASSINFOpdf
\else
\fi
\usepackage{flushend}
\usepackage{url}
\usepackage{algorithm}
\usepackage[noend]{algorithmic}
\usepackage{amssymb}
\usepackage{graphicx}
\usepackage[cmex10]{amsmath}

\newcommand{\rmnum}[1]{\romannumeral #1}

\begin{document}
%
\title{Using Incomplete Information for Complete Weight Annotation of Road Networks--- Extended Version}
%
%
%
%

\author{Bin~Yang$^1$, Manohar~Kaul$^1$, Christian~S.~Jensen$^2$ \\ $^1$Department of Computer Science, Aarhus University, Denmark \\
$^2$Department of Computer Science, Aalborg University, Denmark\\
\{byang, mkaul\}@cs.au.dk, csj@cs.aau.dk}

\IEEEcompsoctitleabstractindextext{%
\begin{abstract}
  We are witnessing increasing interests in the effective use of road
  networks. For example, to enable effective vehicle routing,
  weighted-graph models of transportation networks are used, where the
  weight of an edge captures some cost associated with traversing the
  edge, e.g., greenhouse gas (GHG) emissions or travel time.
  It is a precondition to using a graph model for routing that all
  edges have weights.
  Weights that capture travel times and GHG emissions can be extracted
  from GPS trajectory data collected from the network. However, GPS
  trajectory data typically lack the coverage needed to assign weights
  to all edges.
  This paper formulates and addresses the problem of annotating all
  edges in a road network with travel cost based weights from a set of
  trips in the network that cover only a small fraction of the edges,
  each with an associated ground-truth travel cost.
  A general framework is proposed to solve the problem. Specifically,
  the problem is modeled as a regression problem and solved by
  minimizing a judiciously designed objective function that takes into
  account the topology of the road network.
  In particular, the use of weighted PageRank values of edges is
  explored for assigning appropriate weights to all edges, and the
  property of directional adjacency of edges is also taken into
  account to assign weights.
  Empirical studies with weights capturing travel time and GHG
  emissions on two road networks (Skagen, Denmark, and North Jutland,
  Denmark) offer insight into the design properties of the proposed
  techniques and offer evidence that the techniques are effective.

  This is an extended version of ``Using Incomplete Information for
  Complete Weight Annotation of Road Networks''~\cite{yangtkde}, which is accepted for
  publication in \emph{IEEE TKDE}.
\end{abstract}

\begin{keywords}
Spatial databases and GIS, correlation and regression analysis.
\end{keywords}}

\maketitle

\IEEEdisplaynotcompsoctitleabstractindextext

\IEEEpeerreviewmaketitle

\section{Introduction}

\IEEEPARstart{R}{eduction} in greenhouse gas (GHG) emissions is
crucial in combating global climate change.
For example, the EU has committed to reduce GHG emissions to 20\%
below 1990 levels by 2020~\cite{20reduction}.
To achieve these reductions, the transportation sector needs to
achieve reductions. For example, in the EU, emissions from
transportation account for nearly a quarter of the total GHG
emissions~\cite{emissionTransport}, making transportation the second
largest GHG emitting sector, trailing only the energy sector.

While improved vehicle and engine design are likely to yield GHG
emission reductions, eco-routing is readily deployable and is a
simple yet effective approach to reducing GHG emissions from road
transportation~\cite{guoecomark}.
Specifically, eco-routing can effectively reduce fuel usage and CO$_2$
emissions.  Studies suggest that by providing eco-routes to drivers,
approximately 8--20\% in fuel savings and lower CO$_2$ emissions are
possible in different settings, e.g., during peak versus off-peak
hours, on highways versus areal roads, for light versus heavy duty
vehicles~\cite{kono2008fuel,ericsson2006optimizing}.
For example, an interesting municipal solid waste collection scenario,
where a truck collects solid waste from several locations on Santiago
Island, demonstrates a 12\% fuel reduction due to
eco-routes~\cite{tavares2009optimisation}.

Vehicle routing relies on a weighted-graph representation of the
underlying road network. To achieve effective eco-routing, it is
essential that accurate edge weights that capture environmental costs,
e.g., fuel consumption or GHG emissions, associated with traversing
the edges are available.
Given a graph with appropriate weights, eco-routes can be efficiently
computed by existing routing algorithms, e.g., based on Dijkstra's
algorithm or the $A^{*}$ algorithm.
However, accurate weights that capture environmental impact are not
always readily available for a road network.
This paper addresses the task of obtaining such weights for a road
network from a collection of measured $(\mathit{trip}$,
$\mathit{cost})$ pairs, where the $\mathit{cost}$ can be any cost
associated with a trip, e.g., GHG emissions, fuel consumption, or
travel time.

Because the trips given in the input collection of pairs generally do
not cover all edges of the road network and also do not cover all
times of the day, data sparsity is a key problem.
The cost of a trip, e.g., GHG emissions, differs during peak versus
off-peak hours. Thus, it is inappropriate to use costs associated with
peak-hour trips for obtaining edge weights to be used for eco-routing
during off-peak hours.

\begin{figure}[htp]
\centering
\includegraphics[width=0.90\columnwidth]{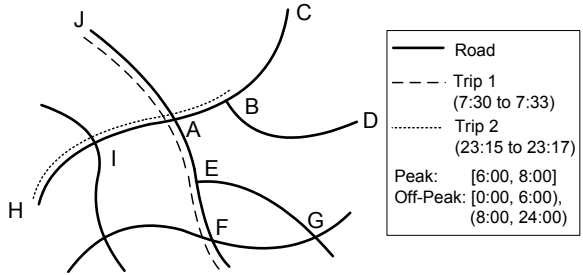}
\caption{Trips on A Road Network}
\label{fig:intro}       
\end{figure}

Considering the road network and trips shown in Fig.~\ref{fig:intro},
assume that the GHG emissions of trip 1 (traversed from 7:30 to
7:33) and trip 2 (traversed from 23:15 to 23:17) are also given,
and assume that we are interested in assigning GHG emission weights to
all edges in the network.
The assignment of these weights to a large number of edges, e.g.,
$BC$, $BD$, $EG$, and $FG$, cannot be done directly since they are not
covered by any trip. However, for example, $BD$ can be annotated by
considering its neighbor road segment $AB$ which is covered by trip~2.

Assuming that the period from 6:00 to 8:00 is the sole peak-hour
period (the remaining times being off-peak), trip~1 is not useful for
assigning an off-peak weight to the edge $AE$ because trip~1 traversed
$AE$ during peak hours. By taking into account the off-peak weights of
$IA$ and $AB$ (covered by trip~2), it is, however, possible to obtain
an off-peak weight for $AE$.

This paper proposes general techniques that take as input (i) a
collection of $(\mathit{trip}$, $\mathit{cost})$ pairs, where
$\mathit{trip}$ captures the edges used and the times when the edges
are traversed and the $\mathit{cost}$ represents the cost of the
entire trip; and (ii) an unweighted graph model of the road network
in which the trips occurred. The techniques then assign travel cost
based weights to all edges in the graph.

To the best of our knowledge, this paper is the first to study
complete weight annotation of road networks using incomplete information. In
particular, the paper makes four contributions.
First, a novel problem, road network weight annotation, is proposed and
formalized.
Second, a general framework for assigning time-varying trip cost based
weights to the edges of the road network is presented, along with
supportive models, including a directed, weighted graph model capable
of capturing time-varying edge weights and a trip cost model based on
time varying edge weights.
Third, two novel and judiciously designed objective functions are
proposed to contend with the data sparsity. A weighted PageRank-based
objective function aims to measure the variance of weights on road
segments with similar traffic flows, and a second objective function aims
to measure the weight difference on road segments that are
directionally adjacent.
Fourth, comprehensive empirical evaluations with real data sets are
conducted to elicit pertinent design properties of the proposed
framework.

The remainder of this paper is organized as follows. Following a
survey of related work in Section~\ref{sec:relatedwork},
Section~\ref{sec:pre} covers problem definition and a general
framework for solving the problem.  Section~\ref{sec:objFunction}
details the objective functions. Section~\ref{sec:exp} reports the
empirical evaluation, and Section~\ref{sec:con} concludes and
discusses research directions.

Compared to the \emph{IEEE TKDE} paper~\cite{yangtkde}, this extended version (i)
provides a mathematical analysis of the foundation of PageRank and
additional justification for the feasibility of using PageRank values
to quantify movement flow based similarities (in
Sections~\ref{ssec:top} and~\ref{ssec:wpc}); (ii) reports on
additional empirical studies aimed at identifying the behavior of
PageRank values on graphs representing the Web, citations, and road
networks; and (iii) provides justification for why the use of PageRank
values are appropriate in the paper's setting (in
Section~\ref{sssec:propPR}).

\section{Related Work}
\label{sec:relatedwork}

Little work has been done on weight annotation of road neworks.
Trip cost estimation is a core component of our weight
annotation solution.
Given a set of $(\mathit{trip}, \mathit{cost})$ pairs as input, trip
cost estimation aims to estimate the costs for trips that do not exist
in the given input set.
Weight annotation can be regarded as a generalized version of trip
cost estimation, since if pertinent weights can be assigned to a road
network, the cost of any trip on the road network can be estimated.
For example, if a GHG emissions based weighted graph is available, the
GHG emissions of a certain trip can be estimated as the sum of the
weights of the road segments that the trip traverses.

Most existing work on trip cost
estimation~\cite{DBLP:conf/aaai/IdeS11, DBLP:conf/sdm/IdeK09,
  clark2003traffic, DBLP:conf/gis/YuanZZXXSH10} focuses on travel-time
estimation. In other words, their work focuses on travel time as the
trip cost.
In general, the methods for estimating the travel times of trips can
be classified into two categories: (\rmnum{1}) segment models and
(\rmnum{2}) trip models.

Segment models~\cite{clark2003traffic,
  DBLP:conf/gis/YuanZZXXSH10,ygnace2000travel,
  herrera2010incorporation} concern travel time estimation for
individual road segments. For example, observers (e.g., Bluetooth
sensors or loop detectors deployed along road segments) monitor the
traffic on road segments, recording the flows of vehicles along the
road segments.
Thus, travel-time estimation tends to concern particular road
segments. For example, some studies model travel time on a particular
road segment as a time series and apply autoregressive
models~\cite{clark2003traffic} to estimate the travel time on the road
segment.
T-Drive~\cite{DBLP:conf/gis/YuanZZXXSH10} models time-dependent travel
time distributions on road segments using sets of histograms and
enables the inference of future travel times using Markov
chains~\cite{DBLP:conf/kdd/YuanZXS11}.
One study incorporates Lagrangian
measurements~\cite{herrera2010incorporation} into existing traffic
flow models for freeways to estimate travel time distributions on
specific freeways.

Segment models assume ``hot'' road segments where, preferably,
substantial data is available. However, far from every road
segment may have enough historical data in practical settings, e.g.,
due to the limited deployment of costly sensors.
Segment models are not well suited for the weight annotation
problem because the given $(\mathit{trip}$, $\mathit{cost})$ pairs
typically fail to cover the whole network, meaning that many road
segments lack the data needed to apply such models.

The trip models focus on estimating the costs of individual trips.
Specifically, the costs of trips are considered more interesting than
the costs of individual road segments.
Given a collection of trips and their corresponding travel times, one
study~\cite{DBLP:conf/sdm/IdeK09} proposes a Gaussian process
regression based method to predict the travel times for unseen trips.
However, the study has the limitation that all the trips are required
to share the same source and target.
This limitation renders the study of limited interest to us, since we
aim at annotating every edge with a pertinent weight.
Trajectory regression~\cite{DBLP:conf/aaai/IdeS11} was proposed
recently to infer the travel times of arbitrary trips. The method is
able to estimate the travel times of trips consisting of road segments
with no or little traversal history by considering the travel time
correlation of spatially adjacent road segments.

Trajectory regression is the most related method to our
weight annotation problem. However, our study distinguishes
itself with several unique characteristics.
First, we propose a general framework for annotating edges in a road
network with a range of trip cost based weights and are not
constrained to travel time.
Second, we identify the cost correlation of road segments sharing
similar traffic flows, and we
quantify this by using weighted PageRank values.
Third, we consider the temporal cost correlation of adjacent road
segments. For example, although two road segments $AB$ and $BC$ are
adjacent, the cost of traversing $AB$ during peak hours is not
necessarily correlated to the cost of traversing $BC$ during
off-peak hours.
Fourth, we take into account the directionality of road segments and
consider only \emph{directional adjacency} when determining the cost
correlation of spatially adjacent road segments.
Last but not least, we conduct comprehensive experiments on real data
sets (real trips and real road networks) to demonstrate the
effectiveness of annotating road networks with both travel time based
weights and GHG emissions based weights.
The earlier study on trajectory
regression~\cite{DBLP:conf/aaai/IdeS11} considers only synthetic data
and estimates only travel times of trips.

In the intelligent transportation system research
field~\cite{guoecomark,song2009aggregate,ahn2002estimating}, other
travel costs (besides travel time) of trips are studied.
For example, fuel consumption and GHG emissions of a trip can be
computed based on instantaneous vehicle velocities and accelerations,
the slopes of the road segments traversed, and the engine type.
However, these methods are designed to estimate the costs of
individual trips and are not readily applicable to the problem of
annotating graph edges with trip cost based weights, notably edges
that do not have any traversed trips.

\section{Preliminaries}
\label{sec:pre}

We cover the modeling that underlies the proposed framework, and we
provide an overview of the framework and its setting.

We use blackboard bold upper case letter for sets, e.g., $\mathbb{E}$,
bold lower case letters for vectors, e.g., $\mathbf{d}$, and bold upper
case letters for matrices, e.g., $\mathbf{M}$.
Unless stated otherwise, the vectors used are column vectors. The
$i$-th element of vector $\mathbf{d}$ is denoted as $\mathbf{d}[i]$,
and the element in the $i$-th row and $j$-th column of matrix
$\mathbf{M}$ is denoted as $\mathbf{M}[i,j]$.
Matrix $\mathbf{M}^\mathbf{T}$ is $\mathbf{M}$ transposed.  An
overview of key notation used in the paper is provided in
Table~\ref{tbl:notation}.
%
\begin{table}[!htbp]
\caption{Key Notation}
\label{tbl:notation}
\small \centering
\begin{tabular}{ll}
\hline\noalign{\smallskip}
Notation & Description   \\
\noalign{\smallskip}\hline\noalign{\smallskip}
$G$, $G'$ & The primal graph and the dual graph.   \\
$G_k'$ & The dual graph in traffic category tag $\mathit{tag_k}.$   \\
$\mathbb{V}$, $\mathbb{E}$ & The vertex set and the edge set.   \\
$\mathbb{V}'$, $\mathbb{E}'$ & The dual vertex set and the dual edge set.   \\
$\mathbf{d}$ & The cost variable vector for all edges.   \\
$\mathit{PR}_{k}(v_i')$ & The weighted PageRank value of dual \\
& vertex $v_i'$ in traffic category tag $\mathit{tag_k}$.   \\
\noalign{\smallskip}\hline
\end{tabular}
\end{table}

\subsection{Modeling a Temporal Road Network}
\label{ssec:modeling}

A road network is modeled as a directed, weighted graph $G =
(\mathbb{V}$, $\mathbb{E}$, $L$, $F$, $H)$, where $\mathbb{V}$ and
$\mathbb{E}$ are the vertex and edge sets, respectively; $L$ is a
function that records the lengths of edges; $F$ is a function that
maps times to traffic categories; and $H$ is a function that
assigns time-varying weights to edges. We proceed to cover each
component in more detail.

A vertex $v_i$ $\in$ $\mathbb{V}$ represents a road intersection or an
end of a road. An edge $e_k \in \mathbb{E} \subseteq \mathbb{V} \times
\mathbb{V}$ is defined by a pair of vertices and represents a directed
road segment that connects the (intersections represented by) two
vertices. For example, edge $(v_i$, $v_j)$ represents a road segments
that enables travel from vertex $v_i$ to vertex $v_j$. For convenience,
we call this graph representation of a road network the \emph{primal
  graph}.

\begin{figure}[!ht]
%
   \begin{minipage}[b]{0.24\textwidth}
     \centering
     \includegraphics[width=\columnwidth]{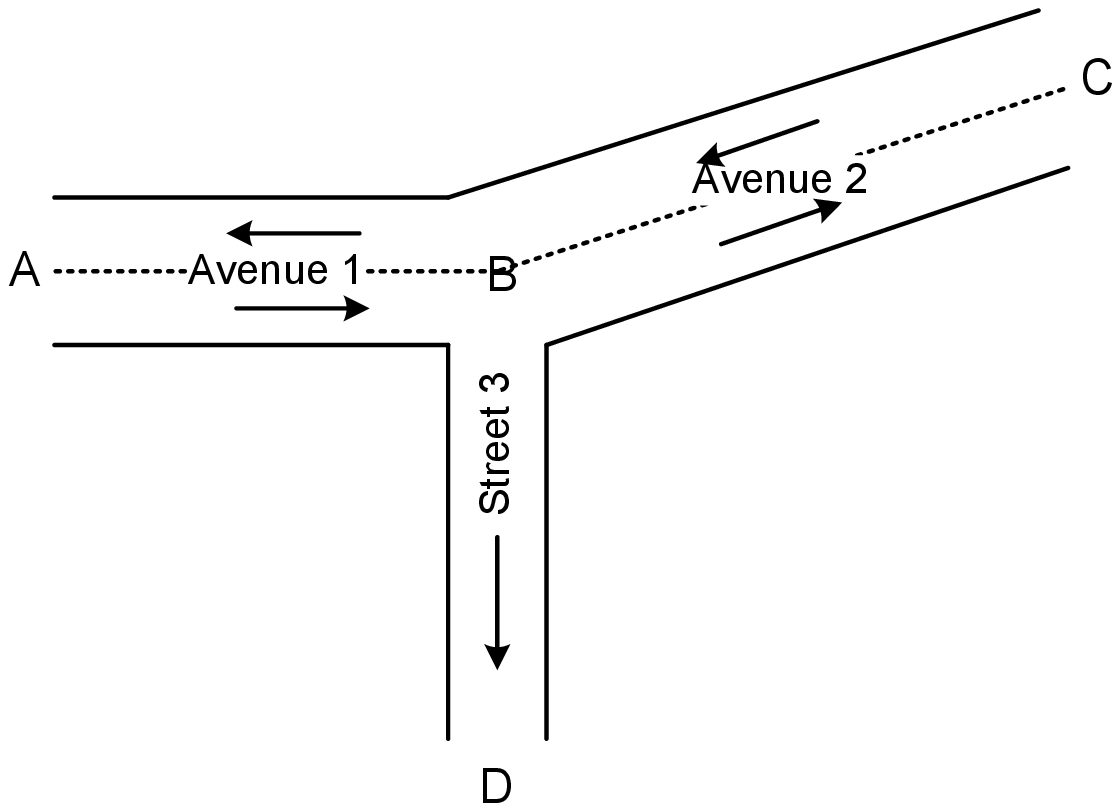}
\caption{Road Network}
     \label{fig:roadnetwork}
   \end{minipage}
   \hfill
   \begin{minipage}[b]{0.24\textwidth}
     \centering
     \includegraphics[width=\columnwidth]{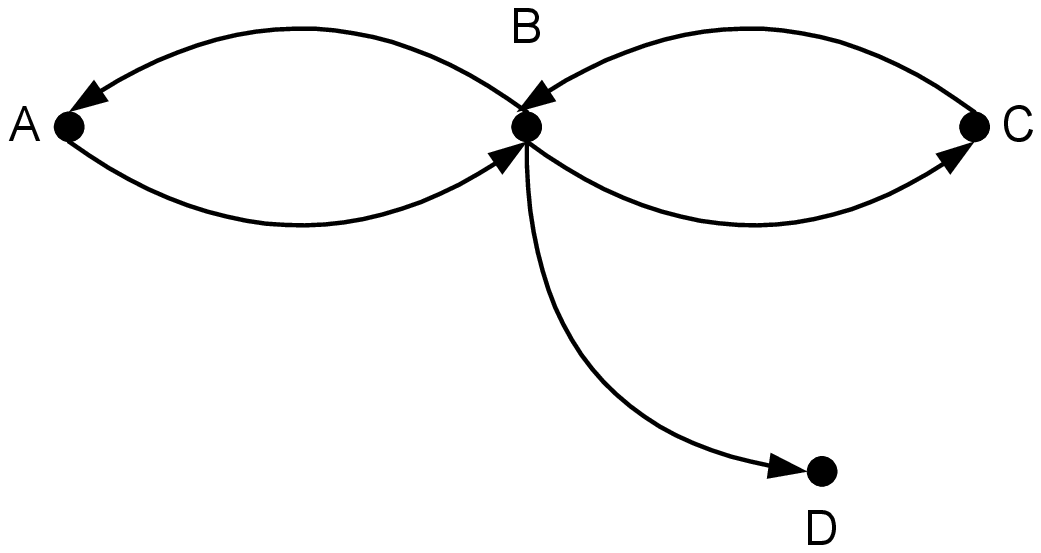}
     \caption{Primal Graph} \label{fig:primal}
   \end{minipage}
\end{figure}

Fig.~\ref{fig:roadnetwork} captures the upper right part of the
road network shown in Fig.~\ref{fig:intro} in more detail.
%
%
%
Here, \emph{Avenue~1} and \emph{Avenue~2} are bidirectional roads, and
\emph{Street~3} is a one-way road that only allows travel from vertex
$B$ to vertex $D$.

The corresponding primal graph is shown in Fig.~\ref{fig:primal}. In
order to capture the bidirectional \emph{Avenue~1}, two edges $(A, B)$
and $(B, A)$ are generated. Since \emph{Street~3} is a one-way road,
only one edge, $(B, D)$, is created.
%
%

It is essential to model a road network as a directed graph because
the cost associated with traveling in two different directions may
differ very substantially.
For example, traveling uphill is likely to have a higher fuel cost
than traveling downhill. As another example, the congestion may
also vary greatly for the two directions of a road.

Function $L: \mathbb{E} \rightarrow \mathbb{R}$ takes as input an edge
and outputs the length of the road segment that the edge represents.
If road segment $AB$ is 135 meters long, we have $G.L((A,B))= G.L((B, A)) =
135$.

Next, the cost of traversing the same edge may differ across
time. This is typically due to varying degrees of congestions. Thus,
GHG emissions or fuel consumption are likely to differ during peak
versus off-peak times.
To this end, function $F: \mathit{TD} \rightarrow \mathit{TAGS}$
models the varying traffic intensity during different periods.
Specifically, $F$ partitions time $\mathit{TD}$ and assigns a
\emph{traffic category tag} in $\mathit{TAGS}$ to each partition.
The granularity of the tags are chosen so that the traffic intensity
can be assumed to be constant during the time associated with the
same tag.
For example, $F([$0:00, 7:00$))=\emph{OFFPEAK}$, $F([$7:00,
9:00$))=\emph{PEAK}$, $F([$9:00, 17:00$))=\emph{OFFPEAK}$,
etc.

Finally, function $H: \mathbb{E} \times \mathit{TAGS} \rightarrow
\mathbb{R}$ assigns time dependent weights to all edges. In
particular, $H$ takes as input an edge and a traffic tag, and outputs
the weight for the edge during the traffic tag.

Specifically, $G.H(e_i$, $tag_j)=d_{(e_i,\ tag_j)}$$\cdot$ $G.L(e_i)$,
where $d_{(e_i,\ tag_j)}$ indicates the cost per unit length of
traversing edge $e_i$ during tag $tag_j$ and $G.L(e_i)$ is the length
of edge $e_i$.
To maintain the different costs on different edges during different
traffic tags, function $H$ maintains
$|E|\cdot$$|\mathit{TAGS}|$ cost variables, denoted as $d_{(e_i,\
  tag_j)}$ (where $1$ $\leqslant$ $i$ $\leqslant|E|$ and $1$
$\leqslant$ $j$ $\leqslant|\mathit{TAGS}|$).

We organize all the cost variables into a \textbf{cost vector}
$\mathbf{d} \in \mathbb{R}^{(|E|\cdot|\mathit{TAGS}|)}$ and $\mathbf{d}$=
$[d_{(e_1,\ tag_1)}$, $\ldots$, $d_{(e_{|E|},\ tag_1)}$, $d_{(e_1,\
tag_2)}$, $\ldots$, $d_{(e_{|E|},\ tag_2)}$, $\ldots$, $d_{(e_1,\
tag_{|\mathit{TAGS}|})}$, $\ldots$, $d_{(e_{|E|},\
tag_{|\mathit{TAGS}|})}$$]^\mathbf{T}$.
The $x$-th element of the vector, i.e., $\mathbf{d}[x]$, equals
$d_{(e_i,\ tag_j)}$ and $x = pos(i, j) = (j-1)$
$\cdot$$|TAGS|+i$.
Note that if the cost vector $\mathbf{d}$ becomes available, the
function $G.H$ also becomes available.

The proposed model is attractive in our setting. It is simpler than
existing models capable of capturing time-varying weights (e.g.,
time-expanded graphs~\cite{DBLP:conf/esa/KohlerLS02} and
time-aggregated graphs~\cite{DBLP:conf/er/GeorgeS06}), and yet it is
sufficiently expressive for the problem we solve.

\subsection{Trips and Trip Costs}
\label{ssec:trip}

Since vehicle tracking using GPS is widespread and growing, we take
into account trips derived from GPS observations.
A GPS trajectory $gpsTr=(gps_1, gps_2, \ldots, gps_n)$ is a
sequence of GPS observations, where a GPS observation $gps_i$
specifies the location of a vehicle at a particular time point.
After map matching and some pre-processing, a GPS trajectory is
transformed into a trip $t=(l_1, l_2, \ldots, l_m)$ that consists of a
sequence of \emph{link records} $l_i$ of the form:
\begin{displaymath}
\begin{small}
\mathit{link\ record}\ l_i:\ \  (e, t_s, t_e),
\end{small}
\end{displaymath}
where $e \in \mathbb{E}$ indicates an edge in $G$ and $t_s$ and
$t_e$ indicate the time points of the first and last GPS
observations on edge $e_i$.

If a graph $G$ is available that contains relevant edge costs, the
cost of a trip $t=(l_1, l_2, \ldots, l_m)$ can be estimated by
Equation~\ref{eq:cost_trip}.
\begin{equation}
\small
\label{eq:cost_trip} \mathit{cost}(t) = \sum_{l_i \in t} \sum_{tag_j
\in \mathit{TAGS}} \mathit{weight}(l_i, tag_j) \cdot G.H(l_i.e,
tag_j),
\end{equation}
where
\[
\begin{small}
\mathit{weight}(l_i, tag_j) =  \frac{\sum_{I \in
G.F^{-1}(tag_j)}| I \cap [l_i.t_s, l_i.t_e]|}{|[l_i.t_s, l_i.t_e]|}.
\end{small}
\]
Here, $G.F^{-1}$ indicates the inverse function of $F$ defined in $G$,
which takes as input a traffic tag and outputs the set of its
corresponding time intervals.
Next, $|\cdot|$ denotes the length of an interval. For example, given
a trip that contains link record $l_i=(e_j, 6:51, 7:05)$ and the
traffic tags given in Section~\ref{ssec:modeling}, the cost of the
trip is $\frac{10}{15}$ $\cdot$ ${G.H}{(e_j, \emph{OFFPEAK})}$ +
$\frac{5}{15}$ $\cdot$ ${G.H}{(e_j, \emph{PEAK})}$ = $\frac{10}{15}$
$\cdot$ ${d}_{(e_j, \emph{OFFPEAK})}$ $\cdot$ $G.L(e_j)$+
$\frac{5}{15}$ $\cdot$ ${d}_{(e_j, \emph{PEAK})}$ $\cdot$ $G.L(e_j)$.

\subsection{Framework Overview}
\label{ssec:problem}

\begin{figure}[htp]
\centering
  \includegraphics[width=0.8\columnwidth]{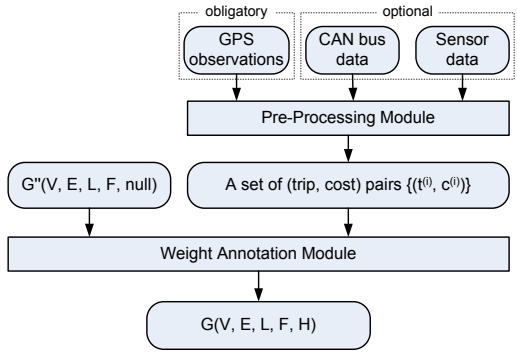}
\caption{Framework Overview}
\label{fig:problem}       
\end{figure}

Fig.~\ref{fig:problem} gives an overview of the framework for
assigning trip cost based weights to a road network.
Various types of raw data collected from a road network, such as
GPS observations with corresponding CAN bus data and sensor data, are
fed into a pre-processing module. While the GPS observations are
obligatory, the CAN bus and sensor data are optional.

\textbf{Pre-processing module}: The GPS observations are map matched
and transformed into trips as defined in Section~\ref{ssec:trip}.
Next, a cost is associated with each trip.  If only GPS observations
are available, some costs, e.g., travel time, can be associated with
trips directly. Other costs, e.g., GHG emissions, can be derived.  For
example, models are available in the literature that are able to
provide an estimate of a trip's GHG emissions and fuel consumption
based on the GPS observations of the trip~\cite{guoecomark}.
If CAN bus data and sensor data are also available along with the GPS
data, actual and more accurate fuel consumption and GHG emissions can
be obtained directly, and thus can be associated with trips.

The pre-processing module outputs a set of $(\mathit{trip}$,
$\mathit{cost})$ pairs $\{(t^{(i)}, c^{(i)})\}$, which then serve as
input to the edge annotation module.
For example, if the goal is to assign GHG emissions based weights,
cost value $c^{(i)}$ indicates the GHG emissions of trip $t^{(i)}$.
Note that the cost $c^{(i)}$ is the total cost associated with the
$i$-th trip, meaning that the cost for each individual link record in
the $i$-th trip is not required to be known. This makes it easier to
collect $(\mathit{trip}$, $\mathit{cost})$ pairs. Because pairs may be
obtained in wide variety of ways, the proposed framework has the
potential for wide applicability.

\textbf{Weight annotation module:} The $(\mathit{trip}$,
$\mathit{cost})$ pairs along with a corresponding un-weighted graph
$G'' = (\mathbb{V}, \mathbb{E}, L, F, \mathit{null})$ are fed into
the weight annotation module. This module assigns pertinent weights
to the edges of the graph, and it outputs an weighted graph $G =
(\mathbb{V}, \mathbb{E}, L, F, H)$.

Recall that function $G.H$ from Section~\ref{ssec:modeling} is defined
by the cost vector $\mathbf{d}$. Given a set of $(\mathit{trip}$,
$\mathit{cost})$ pairs $\mathbb{TC}=\{(t^{(i)}, c^{(i)})\}$, the core
task of this module is to estimate appropriate cost variables in
vector $\mathbf{d}$.
We formulate the weight annotation problem as a supervised learning
problem, namely a regression problem~\cite{bishop2006pattern} that
employs $\mathbb{TC}$ as the training data set to estimate cost
variables in vector $\mathbf{d}$.

The regression problem is solved by minimizing a judiciously designed
objective function composed of three sub-objective items.
The first item measures the misfit between the given actual cost and
the estimated cost (i.e., the cost obtained from the cost model
described in Equation~\ref{eq:cost_trip}) for every trip in
$\mathbb{TC}$.
The second item measures the differences between the cost variables of
two edges whose expected traffic flows (based on topological
structures) are similar.
The third item measures the differences between the cost variables of
two edges which are directionally adjacent.
Further, other appropriate metrics that can quantify the difference
between the cost variables of two edges can also be incorporated into
the module.
Finally, minimizing the objective function is handled by solving a
system of linear equations.

\section{Objective Functions}
\label{sec:objFunction}

Since we regard the problem as a regression problem, we elaborate on
the design of the proposed objective function and the solution to
minimizing the objective function.

\subsection{Residual Sum of Squares}
\label{ssec:objRSS}

In order to obtain an appropriate estimation of the cost vector
$\mathbf{d}$, we need to make sure that for every $(\mathit{trip}$,
$\mathit{cost})$ pair $(t^{(i)}$, $c^{(i)})$ $\in$ $\mathbb{TC}$, the
misfit between the actual cost (e.g., $c^{(i)}$) and the estimated
cost (e.g., $\mathit{cost}(t^{(i)})$ evaluated by
Equation~\ref{eq:cost_trip}, which employs $\mathbf{d}$), is as small
as possible.
To quantify the misfit, the residual sum of squares ($\mathit{RSS}$)
function is applied, where
%
\[
\begin{small}
\label{eq:RSS} \mathit{RSS}(\mathbf{d}) = \sum_{(t^{(i)}, c^{(i)})\in
{\mathbb{TC}}} (c^{(i)} - cost(t^{(i)}))^2.
\end{small}
\]

To facilitate the following discussion, we derive a matrix
representation of the $\mathit{RSS}$ function, as shown in
Equation~\ref{eq:RSS2}.
%
\begin{equation}
\small \label{eq:RSS2}
\mathit{RSS}(\mathbf{d}) = ||\mathbf{c} - \mathbf{Q}^\mathbf{T}
\mathbf{d}  ||_2^2
\end{equation}
Let the cardinality of the set $\mathbb{TC}$ be $N$ (i.e.,
$|\mathbb{TC}|=N$).
We define a vector $\mathbf{c}\in\mathbb{R}^{N}=[c^{(1)}$, $c^{(2)}$,
$\ldots$, $c^{(N)}$$]^\mathbf{T}$, where $c^{(i)}$ is the given actual
cost of the trip $t^{(i)}$, and $(t^{(i)}, c^{(i)})\in\mathbb{TC}$.
A matrix $\mathbf{Q}\in \mathbb{R}^{|\mathbf{d}|\times
  N}=[\mathbf{q^{(1)}}$, $\mathbf{q^{(2)}}$, $\ldots$,
$\mathbf{q^{(N)}}$$]$ is introduced to enable us to rephrase
Equation~\ref{eq:cost_trip} into a matrix representation.
Specifically, $\mathbf{q^{(k)}}$ is the $k$-th column vector in
$\mathbf{Q}$ which corresponds to trip $t^{(k)}$.
If trip $t^{(k)}$ contains a link record $l$ whose corresponding edge
is $e_i$ (i.e., $l.e=e_i$), then
$\mathbf{q^{(k)}}$$[\mathit{pos}(i,j)]=G.L$$(e_i)$ $\cdot$
$\mathit{weight}(l, tag_j)$ where $1\leqslant$ $j\leqslant$
$|\mathit{TAGS}|$; otherwise, it is set to $0$.

Different from ordinary regression problems, minimizing
Equation~\ref{eq:RSS2} is insufficient for determining every cost
variable in $\mathbf{d}$ because the trips in $\mathbb{TC}$ may not
cover all the edges in the road network, e.g., all the edges in
$\mathbb{E}$.
For the edges that are never traversed by any trip in $\mathbb{TC}$,
their corresponding cost variables in $\mathbf{d}$ cannot be
determined by only minimizing the $\mathit{RSS}$ function.

In this case, annotating the edges that do not appear in $\mathbb{TC}$
with weights seems to be difficult and even unsolvable.
In the following, we try to use the topology of the road network to
further propagate and constrain the cost variables in order to assign
an appropriate weight to every edge.

\subsection{Topological Constraint}
\label{ssec:objtop}

The topology of a road network is highly correlated with human
movement flow~\cite{DBLP:journals/gis/Jiang09,crisostomi2011google},
including the movement of both pedestrians and vehicles.
Edges with similar movement flows can be expected to have similar cost
variables.
Thus, if an edge is covered in $\mathbb{TC}$, its cost variable
information can be propagated to the edges that have similar movement
flows.
To this end, we study how to quantify movement flow based similarity
between edges using topological information of road networks.

\subsubsection{Modeling Traffic Flows with PageRank}
\label{ssec:top}

We transfer the idea of using PageRank for the modeling of web surfers
to the modeling of vehicle movement in road networks.
The original
PageRank employs the hyperlink structure of the web to build a
first-order Markov chain, where each web page corresponds to a
state~\cite{DBLP:journals/im/LangvilleM03}. The Markov chain is
governed by a transition probability matrix $\mathbf{M}$.
If web page $i$ has a hyperlink pointing to web page $j$ then $\mathbf{M}[i, j]$
is set to $\frac{1}{\mathit{outDegree}(i)}$; otherwise, it is set to
0. $\mathbf{M}[i, j]$ indicates the probability of transition from
state $i$ to state $j$.
PageRank models a user browsing the web as a Markov process based on
matrix $\mathbf{M}$, and the final PageRank vector is the stationary
distribution vector $\mathbf{x}$ of matrix $\mathbf{M}$. The PageRank
of web page $i$, i.e., $\mathbf{x}[i]$, indicates the probability that
the user visits page $i$ or, equivalently, the fraction of time the
user spends on page $i$ in the long
run~\cite{DBLP:journals/im/LangvilleM03}.

According to the Perron--Frobenius theorem, the existence and
uniqueness of the stationary distribution vector $\mathbf{x}$ can only
be guaranteed if matrix $\mathbf{M}$ is stochastic, non-negative,
irreducible, and primitive~\cite{DBLP:journals/im/LangvilleM03}.
However, the original transition probability matrix $\mathbf{M}$ that
is derived purely from the hyperlink structure of the web cannot be
guaranteed to satisfy these properties, e.g., due to dangling pages
(pages without outlinks). Thus, a series of transformations (adding
outlinks to a virtual page for dangling pages, introducing a damping
factor, etc.) are applied to $\mathbf{M}$ to obtain matrix
$\mathbf{M}'$ that satisfies the above properties. The transformations
are well-known and can be found in the original PageRank
paper~\cite{page1999pagerank} and in classical textbooks on Markov
chains~\cite{grinstead1997introduction}.

The modeling movements of vehicles on a road network as stochastic
processes is well studied in the transportation
field~\cite{daganzo1977stochastic}. In particular, the modeling of
vehicle movements as Markov processes is an easy-to-use and effective
approach~\cite{crisostomi2011google}.
Thus, we build a first-order Markov chain with a transition
probability matrix derived from both the topology of the road network
and the trips that occur in the road network. A state corresponds to
an edge in the primal graph (i.e., a directed road segment), not a
vertex (i.e., a road intersection).

The PageRank value of a state indicates the probability that a vehicle
travels on the edge or, equivalently, the fraction of time a vehicle
spends on the edge in the long run. Thus, the PageRank value is
expected to reflect the traffic flow on the edge.
Further, a series of topological
metrics~\cite{DBLP:journals/gis/Jiang09}, including centrality-based
metrics, small-world metrics, space-syntax metrics, and PageRank
metrics, have been applied to capture human movement flows in urban
environments.
When using a graph representation of an urban environment, it is found
that the classical and weighted PageRank metrics are highly correlated
with human movements~\cite{DBLP:journals/gis/Jiang09,jiang2008self}.
Thus, if two edges have similar PageRank values, the traffic flow on
the two segments should be similar.

When modeling web surfers, PageRank assumes that the Markov chain is
time-homogeneous, meaning that the probability of transferring from
page $i$ to page $j$ has the same fixed value at all times. In other
words, matrix $\mathbf{M}$ is static across time.
In contrast, the time-homogenous assumption does not hold for vehicles
traveling in road networks. For example, during peak hours, the
transition probability from edge $i$ to edge $j$ may be substantially
different from the probability during off-peak hours. Thus, we
maintain a distinct transition probability matrix $\mathbf{M}_k$ for
each traffic category tag $\mathit{tag}_k$. During a particular
traffic tag, we assume the Markov chain to be time-homogeneous.
%
%

\subsubsection{PageRank on Dual Graphs}
\label{ssec:dualgraph}

PageRank was originally proposed to assign prestige to web pages in a
web graph, where web pages are modeled as vertices and the hyper-links
between web pages are modeled as edges.
Unlike the web graph, we are not interested in the prestige of
vertices (i.e., road intersections) in the primal graph
representation of a road network; rather, we are interested in the
prestige of edges (i.e., directed road segments).

In order to assign PageRank values to edges, the primal graph
$G=(\mathbb{V}$, $\mathbb{E}$, $L$, $F$, $H)$ is transformed into a
dual graph $G'=(\mathbb{V}'$, $\mathbb{E}')$, where each vertex in
$\mathbb{V}'$ corresponds to an edge in the primal graph, and where
each edge in $\mathbb{E}'$, denoted by a pair of vertices in
$\mathbb{V}'$, corresponds to a vertex in the primal graph.
Since functions $L$, $F$, and $H$ are not of interest in this
section, we do not keep them in the dual graph.

To avoid ambiguity, we use the terms edge and vertex when referring
to primal graphs and use \emph{dual edge} and \emph{dual vertex}
when referring to dual graphs.
Further, we use the term weight when referring to the weight of an edge
in a primal graph, and we use \emph{dual weight} in the context of
dual edges in a dual graph. 

\begin{figure}[!hpt]
\centering
  \includegraphics[width=0.4\columnwidth]{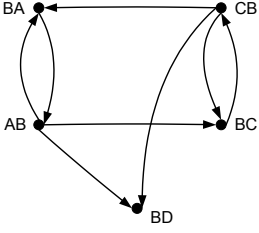}
\caption{Dual Graph}
\label{fig:dual}       
\end{figure}

We define a mapping $\mathit{D2P}: \mathbb{V}' \cup \mathbb{E}'
\rightarrow \mathbb{V} \cup \mathbb{E}$ to record the correspondence
between the elements in the dual and primal graphs.
Fig.~\ref{fig:dual} show the dual graph that corresponds to the primal
graph shown in Fig.~\ref{fig:primal}.
Since the dual vertex $AB$ corresponds to the edge $(A, B)$ in
Fig.~\ref{fig:primal}, $\mathit{D2P}$($AB)=(A, B)$.
Similarly, since the dual edge $(CB, BA)$ corresponds to the vertex $B$ in
Fig.~\ref{fig:primal}, $\mathit{D2P}$$((CB, BA))=B$.

The dual graph is able to model an important characteristic of a road
network: at a particular intersection, the probability of which
segment a vehicle follows depends on the segment via which the vehicle
entered the intersection.
Considering the road network shown in Fig~\ref{fig:roadnetwork}, at
intersection (i.e., vertex) $B$, a vehicle can proceed to follow
segments (i.e., edges) $(B, A)$, $(B, C)$, or $(B, D)$. If a vehicle
entered the intersection using segment $(C, B)$, it may be unlikely
that the vehicle takes a u-turn to follow segment $(B, C)$, while is
more likely that it will use the other segments. Similar cases exist
if a vehicle arrived at the intersection using segment $(A, B)$.

Modeling this characteristic in a primal graph is not easy. For
example, we need to maintain two sets of probabilities on edge $(B,
C)$, for the vehicles came from edge $(C, B)$ versus edge $(A, B)$.
In contrast, modeling this in a dual graph is straightforward, as how
a vehicle entered a particular intersection is clearly represented as
a dual vertex. For example, the probabilities on dual edges $(CB, BC)$
and $(AB, BC)$ record the probabilities that a vehicle entered
intersection $B$ from edge $(C, B)$ and edge $(A, B)$, respectively,
and continues along edge $(B, C)$.

Given the dual graph $G'=(\mathbb{V}'$, $\mathbb{E}')$, original
PageRank values are defined formally as follows.
\begin{equation}
\small \label{eq:PRoriginal}
\mathit{PR}(v_i') = \frac{1-df}{|\mathbb{V}'|} + df
\cdot \sum_{v_j' \in \mathit{IN}(v_i')}
\frac{\mathit{PR}(v_j')}{|\mathit{OUT}(v_j')|}, \ \  v_i' \in \mathbb{V}',
\end{equation}
where $\mathit{PR}(v_i')$ indicates the PageRank value of dual vertex
$v_i'$; $\mathit{IN}(v_i')$ indicates the set of in-link neighbors of
$v_i'$, i.e., $\mathit{IN}(v_i')$ = $\{v_x'$$|(v_x'$, $v_i')$ $\in$
$\mathbb{E}'\}$; and $\mathit{OUT}(v_j')$ indicates the set of
out-link neighbors of $v_j'$, i.e., $\mathit{OUT}(v_j')$ =
$\{v_x'$$|(v_j'$, $v_x')$ $\in$ $\mathbb{E}'\}$. Further, $df$ $\in$
$[0,1]$ is a damping factor, which is normally set to 0.85 for ranking
a web graph.

The intuition behind Equation~\ref{eq:PRoriginal} is that the PageRank
values are composed of two parts: jumping to another random vertex and
continuing the random walk.
This assumption works fine on the web graph, but we need to adapt this
to the different characteristics of the graph representing a road
network.
In a road network, it is impossible for a vehicle to choose a random
edge to traverse when at an intersection. Rather, it can only choose
to continue along one of the out-link (dual) edges.
Based on this observation, we set the damping factor $df$ to $1$. Some
existing empirical studies~\cite{DBLP:journals/gis/Jiang09} also
suggest that with the damping factor set to $1$, the resulting
PageRank values have the best correlation with the human movement
flows.

\subsubsection{Weighted PageRank Computation}
\label{ssec:wpc}

$\\$ \textbf{Definition of Dual Weights:} In the original PageRank
algorithm, a vertex propagates its PageRank value evenly to all its
out-link neighbors. In other words, the dual weight for each dual edge
from dual vertex $v_j'$ is set uniformly to
$\frac{1}{|\mathit{OUT}(v_j')|}$.
The uniform weights on the web graph indicate that a web surfer
chooses its next target web page without any preferences to continue
its random surfing.
However, in a road network, such non-preference surfing usually does
not occur. For example, the next step where a vehicle continues often
depends on where the vehicle came from, as discussed in
Section~\ref{ssec:dualgraph}.
Also, if \emph{Avenue 1} and \emph{Avenue 2} are the main roads in
the road network shown in Fig.~\ref{fig:roadnetwork}, more vehicles
travel from $AB$ to $BC$ than from $AB$ to $BD$.
Further, during different traffic category tags, the transitions
between dual vertices may also be quite different.

With the availability of very large collections of GPS data, we are
able to capture the probability that a vehicle transits from one road
segment to another at an intersection during different traffic
category tags.
Assume we only distinguish between peak and off-peak hours, i.e.,
there are only two corresponding tags in $\mathit{TAGS}$. Suppose we
obtain the number of trips that occurred on the dual edges, as shown
in Table~\ref{table:histrips}.
\begin{table}[!htp]
\caption{Numbers of Trips on Dual Edges }
\label{table:histrips}
\small \centering
\begin{tabular}{lccc}
\hline\noalign{\smallskip}
 Tags & $(AB, BC)$ &  $(AB, BD)$  & $(AB, BA)$ \\
\noalign{\smallskip}\hline\noalign{\smallskip}
 \emph{PEAK} & 30 & 10 & 0  \\
 \emph{OFFPEAK} & 5 & 5   & 0   \\
\noalign{\smallskip}\hline
\end{tabular}
\end{table}

For example, among all the trips that occurred on dual vertex $AB$
during the peak hours, $30$ trips proceeded to follow $BC$, and $10$
trips followed $BD$; during off-peak hours, $5$ trips followed $BC$,
and $5$ trips followed $BD$.
These observations suggest that the dual weight on dual edge $(AB$,
$BC)$ should be greater than the dual weight on dual edge $(AB$, $BD)$
during peak hours; while they should be the same during off-peak
hours.

As the dual graph has different dual weights for different traffic
tags, we need to maintain a dual graph for each traffic tag.
Specifically, the training data set $\mathbb{TC}$ is partitioned into
$\mathbb{TC}_{1}$, $\mathbb{TC}_{2}$, $\ldots$,
$\mathbb{TC}_{|\mathit{TAGS}|}$ according to the traversal times.
Partition $\mathbb{TC}_{k}$ consists only of the trips that
occurred during the time period indicated by the traffic tag $tag_k$,
i.e., $G.F^{-1}(tag_k)$.

The dual weight of a dual edge $(v_i', v_j')$ during tag $tag_k$ is
related to the ratio of the number of trips that traversed the dual
vertices $v_i'$ and $v_j'$ to the number of trips that traversed the
dual vertex $v_i'$, during tag $tag_k$.
Further, to contend with data sparsity, Laplace smoothing is applied
to smooth the dual weight values for the dual edges that are not
covered by any trip in $\mathbb{TC}$.
The dual weight of dual edge $(v_i'$, $v_j')$ for the dual graph
within $tag_k$ (denoted as $G'_{k}$) is computed based on
Equation~\ref{eq:weightTime}.
\begin{equation}
\small \label{eq:weightTime}
W_k(v_i', v_j') = \frac{|\mathit{Trip_k}(v_i',
v_j')|+1} {\sum_{v_x' \in \mathit{OUT(v_i')}} |\mathit{Trip_k}(v_i',
v_x')|+|\mathit{OUT}(v_i')|},
\end{equation}
where $\mathit{Trip_k}(v_i', v_j')$ returns the set of trips in
partition $\mathbb{TC}_k$ that traversed the dual vertices $v_i'$
and $v_j'$.

Continuing the example shown in Table~\ref{table:histrips}, although
no trip goes from the dual vertex $AB$ directly back to $BA$ in
$\mathbb{TC}$, this does not mean that such a trip will not occur in
the future.  Thus, we need to give a small, non-zero value to the dual
weight of dual edge $(AB$, $BA)$.
Using the dual weights provided by Equation~\ref{eq:weightTime}, the
dual weights of the out-linking dual edges of dual vertex $AB$ are:
$W_{\mathit{PEAK}}(AB$, $BC)=\frac{31}{43}$, $W_{\mathit{PEAK}}(AB$,
$BD)=\frac{11}{43}$, and $W_{\mathit{PEAK}}(AB$, $BA)=\frac{1}{43}$;
and $W_{\mathit{OFFPEAK}}(AB$, $BC)=\frac{6}{13}$,
$W_{\mathit{OFFPEAK}}(AB$, $BD)=\frac{6}{13}$, and
$W_{\mathit{OFFPEAK}}(AB$, $BA)=\frac{1}{13}$.

Note that for a given dual vertex $v_i'$, if no trips in $\mathbb{TC}$
are available to assign the dual weights during a traffic tag
$\mathit{tag}_k$, i.e., $|\mathit{Trip}_k(v_i', v_x')|$ = 0 for every
$v_x'$ $\in$ $\mathit{OUT}$$(v_i')$,
Equation~\ref{eq:weightTime} assigns weights with
$\frac{1}{|\mathit{OUT}(v_i')|}$ to each dual edge, which is exactly
what the original PageRank algorithm does.
For instance, if no trips are available for dual vertex $AB$ (i.e., if
the numbers in Table~\ref{table:histrips} are all zeros), the dual
weights for $W_k(AB$, $BC)$, $W_k(AB$, $BD)$, and $W_k(AB$, $BA)$ are
all $\frac{1}{3}$.

Finally, we show why the constructed matrix $\mathbf{M}_k$ is
stochastic, non-negative, irreducible, and primitive, thus ensuring
that the PageRank vector exists uniquely, i.e., convergence is
guaranteed.
Equation~\ref{eq:weightTime} guarantees that the sum of the elements
in a row in matrix $\mathbf{M}_k$ is 1, meaning that $\mathbf{M}_k$ is
stochastic (in particular, stochastic by rows).
Equation~\ref{eq:weightTime} also guarantees that all elements in
matrix $\mathbf{M}_k$ are non-negative.

Irreducibility means that it is possible to each every vertex from
every vertex~\cite{grinstead1997introduction}. This is also true for a
graph representing a road network, where a vehicle can go from every
road segment to every road segment.
A matrix is primitive if some power of the matrix has only positive
elements. Intuitively, this means that for some $n$, it is possible to
go from any vertex to another vertex in $n$ steps. This is not always
guaranteed for a matrix representing a road network. However, simple
mathematical operations can transfer a non-primitive matrix to a
primitive matrix, and the two matrices have the same stationary
distribution vector, i.e., the same PageRank
vector~\cite{grinstead1997introduction}.
Since such mathematical operations are normally implemented in various
packages for computing PageRank values, a stochastic, non-negative,
and irreducible $\mathbf{M}_k$ is sufficient to guarantee convergence
of the PageRank vector on $\mathbf{M}_k$.

\noindent
\textbf{Computing Weighted PageRank Values:} Based on the dual weights
obtained from Equation~\ref{eq:weightTime}, we construct the
transition probability matrices
$\mathbf{M_k}$$\in$$\mathbb{R}^{|\mathbb{V}'|\times|\mathbb{V}'|}$.
Specifically, the $i$th row and $j$th column element in
$\mathbf{M_k}$, i.e., $\mathbf{M_k}$$[i$, $j]$, equals $W_k(v_i',
v_j')$ if the dual edge $(v_i', v_j')$ exists in the dual graph;
otherwise, it equals $0$.
Note that the sum of all elements in a row equals $1$, i.e.,
$\sum_{j=1}^{|\mathbb{V}'|}$$\mathbf{M_k}$$[i$, $j]=1$ for every
$1\leqslant i\leqslant |\mathbb{V}'|$.

Let vector $\mathbf{v_k}$$\in$$\mathbb{R}^{|\mathbb{V}'|}$ record the
PageRank values for every dual vertex in $G'_k$.  Specifically,
$\mathbf{v_k}$$[i]=\mathit{PR_k}$$(v_i')$, which is the PageRank
value of $v_i'$ during traffic category tag $\mathit{tag}_k$.
This way, the PageRank values can be computed iteratively as follows
until converged.
\[
\begin{small}
\label{eq:PRMatrix}
\mathbf{v_k}^{(n+1)} = \mathbf{M_k}^{\mathbf{T}} \cdot \mathbf{v_k}^{(n)},
\end{small}
\]
where $\mathbf{v_k}^{(n)}$ is the PageRank vector in the $n$-th
iteration.

\subsubsection{PageRank-Based Topological Constraint Objective Function}
\label{sssec:PRTC}

After obtaining the weighted PageRank values for every dual edge, the
topological similarity between two edges in the primal graph is
quantified in Equation~\ref{eq:topsim}.
\begin{equation}
\small
\label{eq:topsim} S_{k}^{\mathit{PR}}(e_i, e_j) =
\frac{\min(\mathit{PR}_k(v_{e_i}'),
\mathit{PR}_k(v_{e_j}'))}{\max(\mathit{PR}_k(v_{e_i}'),
\mathit{PR}_k(v_{e_j}'))}
\end{equation}

The topological similarity between edges $e_i$ and $e_j$, denoted as
$S_{k}^{\mathit{PR}}(e_i, e_j)$, is defined based on the weighted
PageRank values of the two dual vertices representing the edges.
To be specific, $v_{e_i}'$ and $v_{e_j}'$ indicate the corresponding
dual vertices of edges $e_i$ and $e_j$, i.e.,
$\mathit{D2P}$$(v_{e_i}')=e_i$ and $\mathit{D2P}$$(v_{e_j}')=e_j$.
Note that Equation~\ref{eq:topsim} returns a high similarity if two
edges have similar weighted PageRank scores and that it returns a low
similarity, otherwise.

Based on the topological similarity, a PageRank-based Topological
Constraint ($\mathit{PRTC}$) function is incorporated into the overall
objective function.
The intuition behind the $\mathit{PRTC}$ function is that for the
same traffic category tag, if two edges have similar traffic flows (as
measured by Equation~\ref{eq:topsim}), their cost variables tend to be
similar as well.
The $\mathit{PRTC}$ function is defined in Equation~\ref{eq:PRTC}.
\begin{equation}
\small \label{eq:PRTC} \mathit{PRTC}(\mathbf{d}) =
\sum_{k=1}^{|\mathit{TAGS}|} \mathit{PRTC}(\mathbf{d}, k),
\end{equation}
where
%
\begin{displaymath}
\begin{small}
\mathit{PRTC}(\mathbf{d}, k) = \sum_{i,j=1}^{|G.\mathbb{E}|}
S_{k}^{PR}(e_i, e_j) \cdot (d_{(e_i, tag_k)}-d_{(e_j, tag_k)})^2.
\end{small}
\end{displaymath}

The value of the $\mathit{PRTC}$ function over the cost vector
$\mathbf{d}$ is the sum of $\mathit{PRTC}(\mathbf{d}, k)$ for every
$1$ $\leqslant$ $k$ $\leqslant |\mathit{TAGS}|$.
The function $\mathit{PRTC}(\mathbf{d}, k)$ computes the weighted
(decided by $S_{k}^{\mathit{PR}}$) sum of the squared differences of
between each pair of road segments' cost variables during traffic tag
$tag_k$.

The $\mathit{PRTC}$ function has two important features: (\rmnum{1})
if the PageRank values of two edges are similar, the similarity value
$S_k^{\mathit{PR}}$ is large, thus making the difference between their
cost variables obvious;
(\rmnum{2}) if two edges' PageRank values are dissimilar, the
similarity value $S_k^{\mathit{PR}}$ with a small value smoothes down
the difference between their cost variables.
This way, minimizing the $\mathit{PRTC}$ function corresponds to
minimizing the overall difference between two cost variables whose
corresponding road segments have similar traffic flows.

To obtain the matrix representation of the $\mathit{PRTC}$ function,
we introduce a matrix $\mathbf{A}\in \mathbb{R}^{|\mathbf{d}|\times
  |\mathbf{d}|}$, which is a block diagonal matrix.
\begin{equation}
\small \label{eq:A}
 \mathbf{A} = \begin{bmatrix}
       \mathbf{A_1}    &  &    &          \\
         & \mathbf{A_2} &   &              \\
               &  & \ldots &           \\
               &   &   & \mathbf{A_{|\mathit{TAGS}|}}
     \end{bmatrix}
\end{equation}
where $\mathbf{A_k}\in \mathbb{R}^{|E|\times |E|}$ and
$\mathbf{A_k}[i$, $j]=S_{k}^{\mathit{PR}}(e_i$, $e_j)$, which
obviously is a symmetric matrix.
Let matrix $\mathbf{L_A}$ be the graph Laplacian induced by the
similarity matrix $\mathbf{A}$. Specifically, $\mathbf{L_A}$$[i$, $j]$
= $\delta_{i,j}$$\cdot$$\sum_{x}$$\mathbf{A}[i,x]-\mathbf{A}[i,j]$,
where $\delta_{i, j}$ returns 1 if $i$ equals $j$, and 0 otherwise.
The matrix representation of $\mathit{PRTC}$ function is shown in
Equation~\ref{eq:PRTCMatrix}.
%
\begin{equation}
\small \label{eq:PRTCMatrix} \mathit{PRTC}(\mathbf{d}) =
\mathbf{d}^\mathbf{T}\mathbf{L_A} \mathbf{d}
\end{equation}

\subsubsection{Properties of PageRank on Road Networks}
\label{sssec:propPR}

Web graphs, citation graphs, and road network graphs are quite
different, rendering it of interest to study the distributions of
PageRank values on these kinds of graphs.
We consider three directed graphs, \emph{WEB}, \emph{CIT}, and
\emph{NJ}, representing a part of the web, citations in a particular
domain, and a road network, respectively.
\begin{enumerate}
\item
  \emph{WEB}\footnote{\url{http://snap.stanford.edu/data/web-Google.html}}:
  Vertices represent web pages, and directed edges represent
  hyperlinks between them. This dataset was released as a part of the
  2002 Google programming contest.
\item
  \emph{CIT}\footnote{\url{http://snap.stanford.edu/data/cit-HepTh.html}}:
  Vertices represent papers, and directed edges represent citations
  between them. This dataset is obtained from the arXiv e-print
  archive in the domain of high-energy physics theory. This dataset
  was released as a part of the 2003 KDD CUP.
\item \emph{NJ}: Vertices represent road segments, and directed edges
  represent road junctions that enable movements between road
  segments. This dataset is the dual graph representation of the North
  Jutland, Denmark road network used in this paper.
\end{enumerate}

\noindent
\textbf{Basic Statistics of The Graphs:} Basic statistics, including
the numbers of vertices (\# V) and edges (\# E), the maximum in-degree
(MI) and out-degree (MO), and the average degree (AD) of the three
graphs are shown in Table~\ref{tbl:data}.
\begin{table}[h]
\caption{Dataset Statistics}
\label{tbl:data}
\centering
\begin{tabular}{|l|r|r|r|r|r|}
  \hline
   & \# V~~ & \# E~~~ & MI~~& MO~ & AD~~\\ \hline
  \emph{WEB} & 875,713 & 5,105,039 & 6,326 &  456 & 5.83\\ \hline
  \emph{CIT} & 27,770 & 352,807 & 2,414 &  562 & 12.70 \\ \hline
  \emph{NJ} & 39,372 & 104,880& 6 & 6 & 2.66 \\ \hline
\end{tabular}
\end{table}

\begin{figure*}[!htp]
\centering
\begin{tabular}{@{}c@{}c@{}c@{}}
\includegraphics[width=0.30\textwidth]{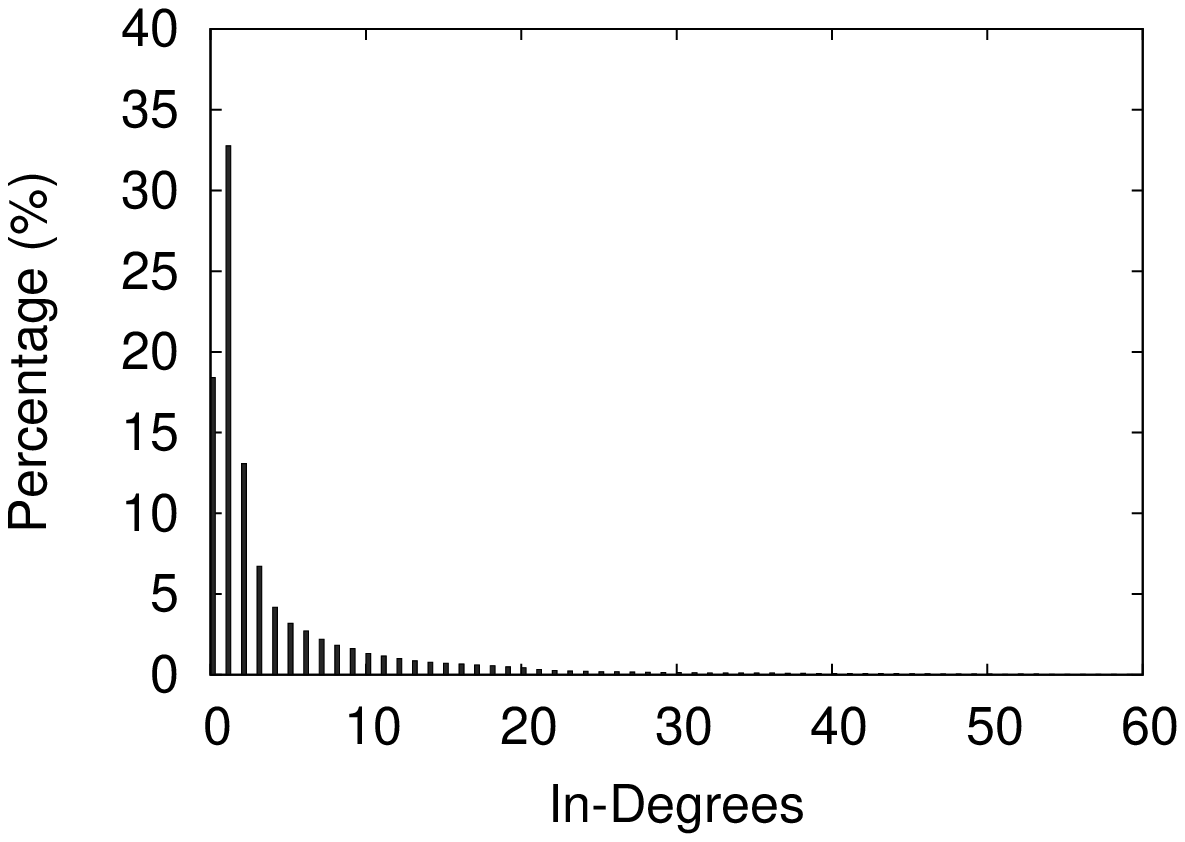}
    &
\includegraphics[width=0.30\textwidth]{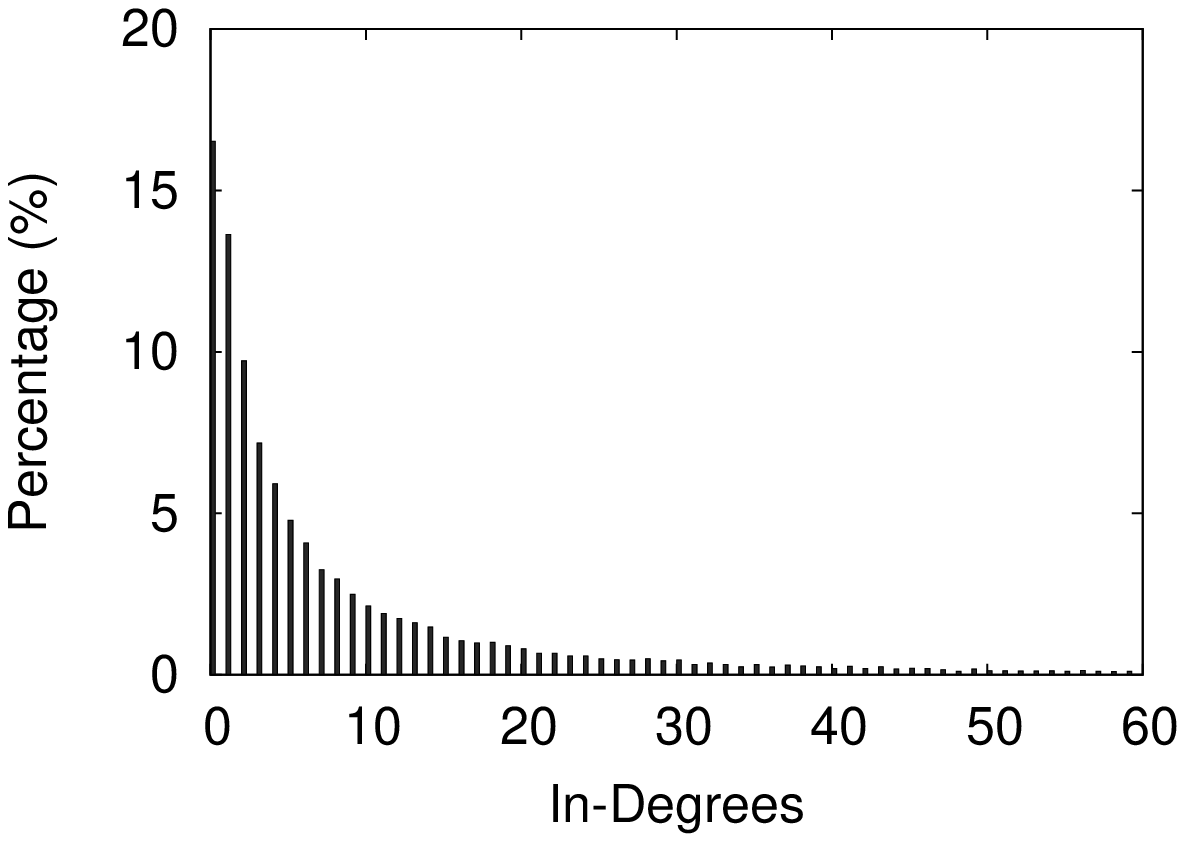}
    &
\includegraphics[width=0.30\textwidth]{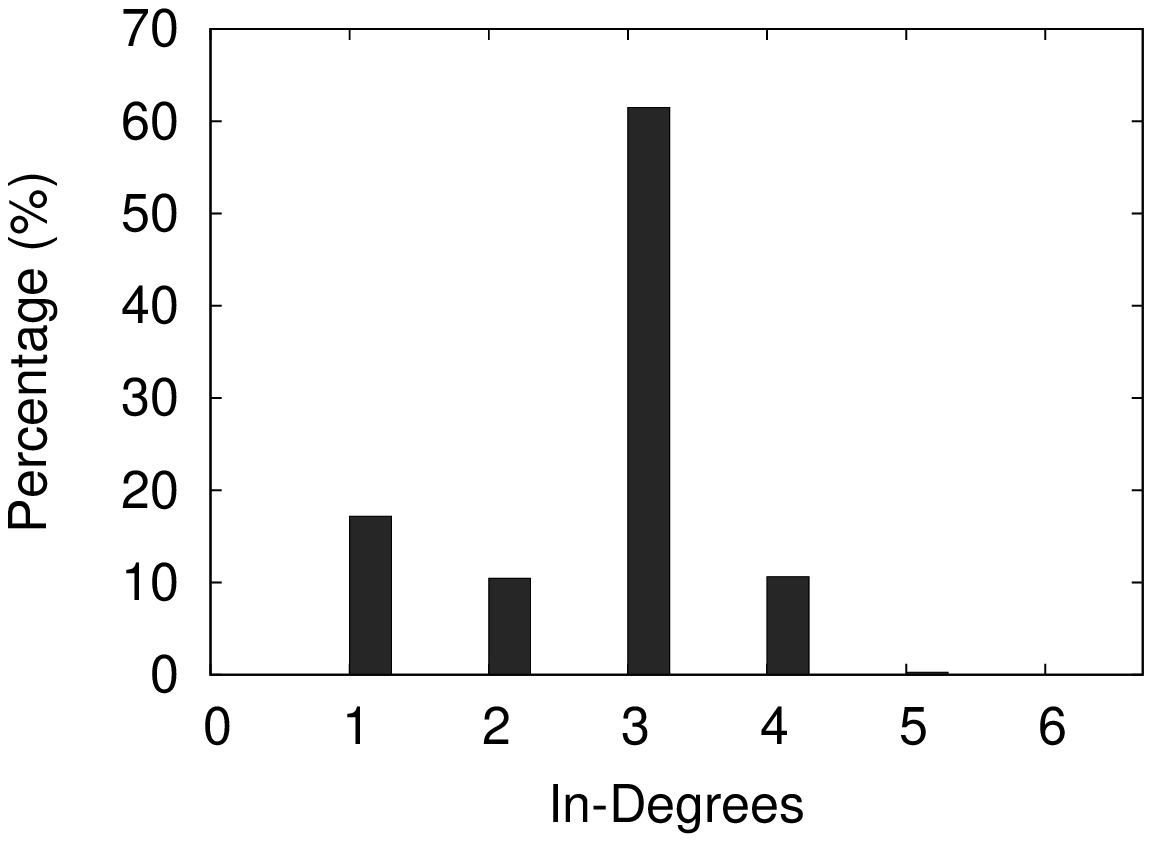}
    \\
    (a) \emph{WEB}
    &
    (b) \emph{CIT}
    &
    (c) \emph{NJ}
\end{tabular}
\caption{In-Degree Distribution}
\label{fig:indegree}
\end{figure*}

\begin{figure*}[!htp]
\centering
\begin{tabular}{@{}c@{}c@{}c@{}}
\includegraphics[width=0.30\textwidth]{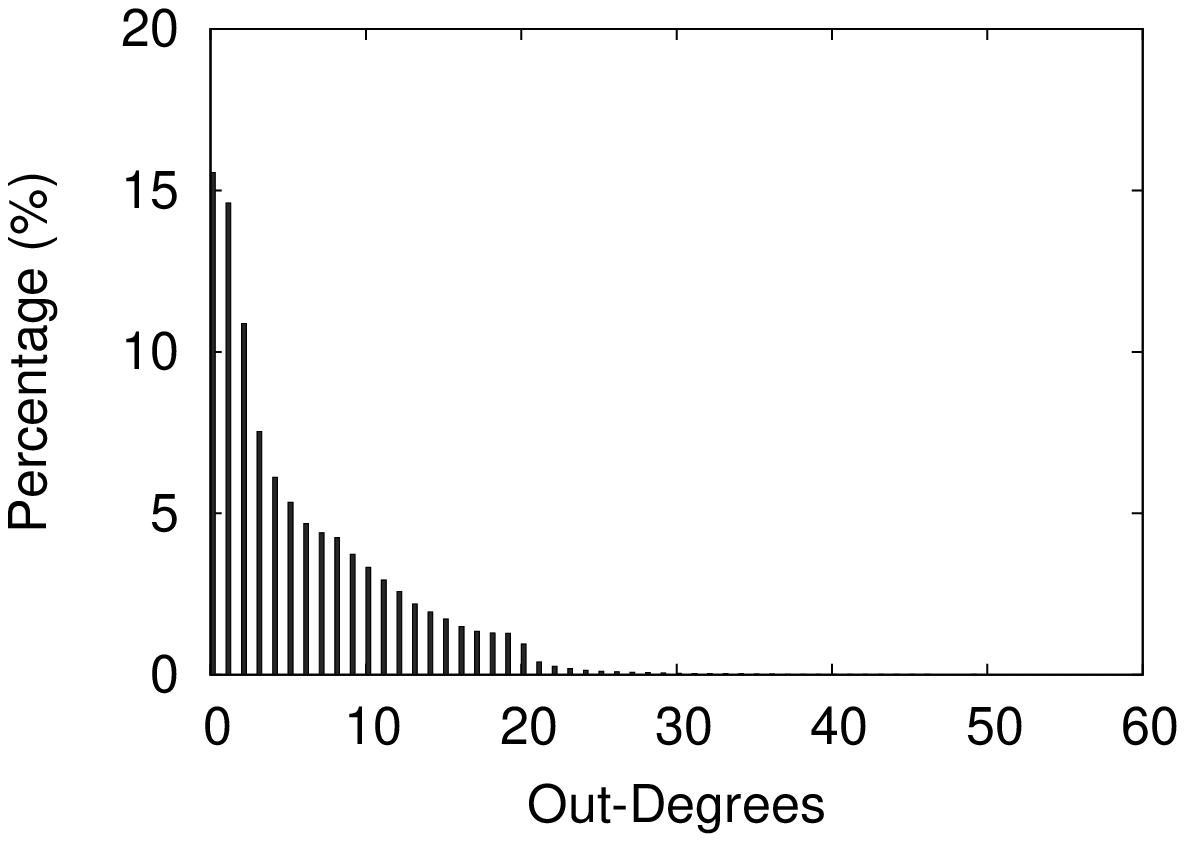}
    &
\includegraphics[width=0.30\textwidth]{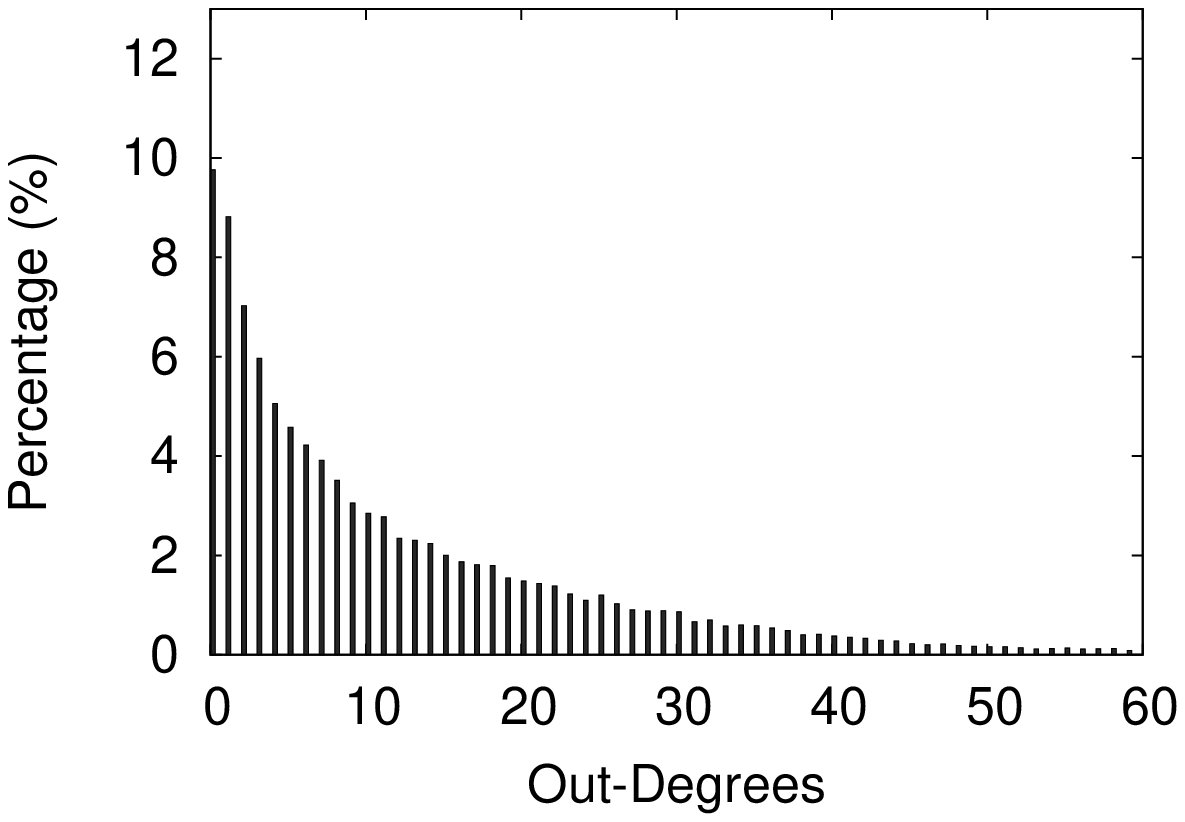}
    &
\includegraphics[width=0.30\textwidth]{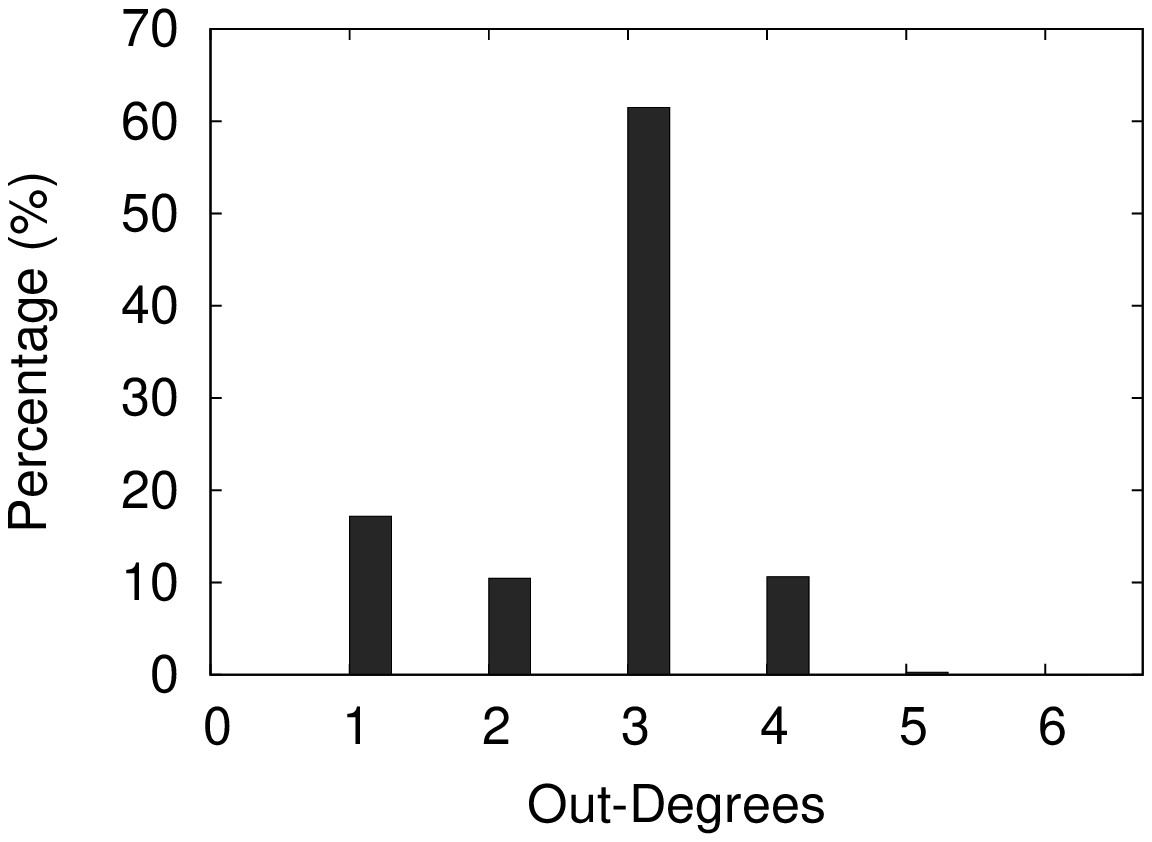}
    \\
    (a) \emph{WEB}
    &
    (b) \emph{CIT}
    &
    (c) \emph{NJ}
\end{tabular}
\caption{Out-Degree Distribution}
\label{fig:Outdegree}
\end{figure*}

To understand better how in-degrees and out-degrees distribute among
vertices, Figures~\ref{fig:indegree} and~\ref{fig:Outdegree} show
in-degrees and out-degrees on the x-axis and the corresponding
percentages of vertices on the y-axis. For \emph{WEB} and \emph{CIT},
Figures~\ref{fig:indegree} and~\ref{fig:Outdegree} only show the
percentage up to degree 60---the percentages for higher degrees are so
small that they are invisible.

All the statistics show that \emph{NJ} is quite different from
\emph{WEB} and \emph{CIT}.
For \emph{WEB} and \emph{CIT}, the degrees (both in-degrees and
out-degrees) vary a lot, from 0 to hundreds or even thousands. The
distribution of degrees are also quite biased: most vertices have very
small degrees. Taking \emph{WEB} as an example, 64.2\%, 78.3\%, and
87.8\% of the vertices have in-degrees smaller than 2, 5, and 10,
respectively.
For \emph{NJ}, the degrees (both in-degrees and out-degrees) of
vertices vary only little, and only from 1 to 6, with most vertices
having degree 3 and almost no vertices having degrees above 4.

Since the basic statistics of \emph{NJ} are quite different from those
of \emph{WEB} and \emph{CIT}, we expect the distribution of PageRank
values for \emph{NJ} to also be different.

\noindent
\textbf{Statistics of the PageRank Values on the Graphs:} As discussed
in the coverage of the mathematical foundation of PageRank, PageRank
is the stationary distribution vector $\mathbf{x}$ of the transition
matrix, which satisfies the property $\sum_{i=1}^{N}\mathbf{x}[i]=1$
($N$ is the number of vertices in a graph).
Since the three graphs have different numbers of vertices, a direct
comparison of absolute PageRank values is meaningless. For instance,
for a graph with 10 vertices, the average PageRank is $\frac{1}{10}$,
whereas for a graph with 1000 vertices, the average PageRank is
$\frac{1}{1000}$. Thus, for example, the PageRank value of a vertex in
the first graph is likely to be larger than that of a vertex in the
second graph.

To conduct a meaningful analysis on the distribution of PageRank
values, we normalize the original PageRank values into range (0, 100].
Given a PageRank vector $\mathbf{x}$, let the $k$-th ($k = \arg
\max_{i\in\{1, 2, \ldots, N\}} \mathbf{x}[i]$) element in the vector
have the biggest PageRank value.
The normalized PageRank vector $\mathbf{y}$ can be computed by
$\mathbf{y}[i]=\frac{\mathbf{x}[i]}{\mathbf{x}[k]} \cdot 100$.

\begin{figure*}[!htp]
\centering
\begin{tabular}{@{}c@{}c@{}c@{}}
\includegraphics[width=0.30\textwidth]{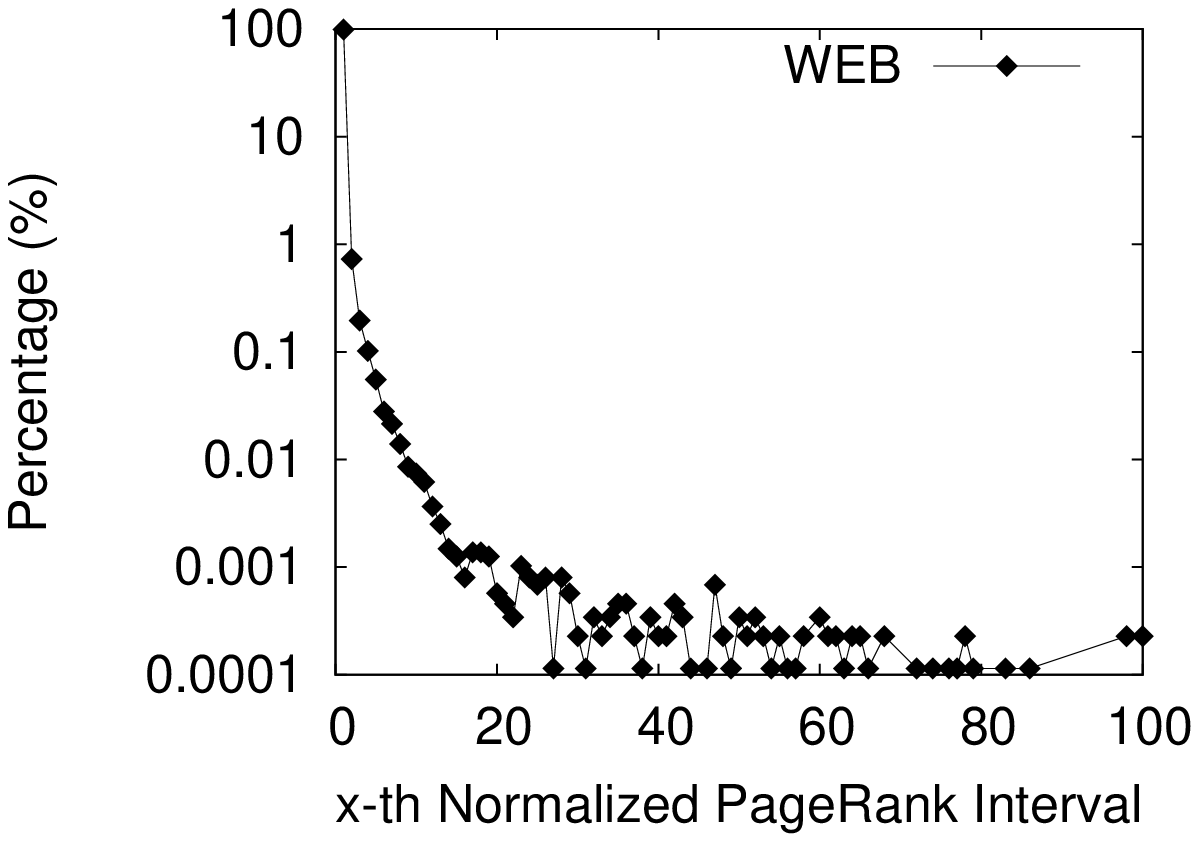}
    &
\includegraphics[width=0.30\textwidth]{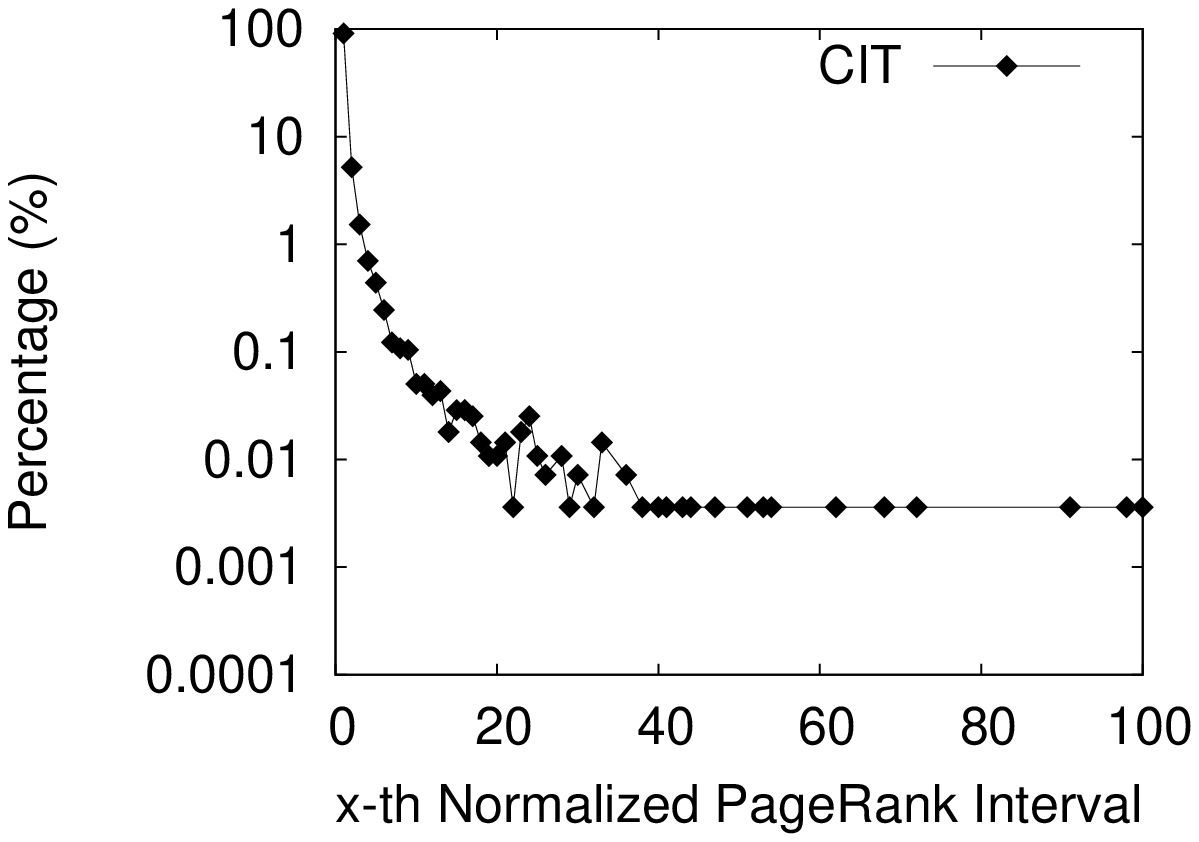}
    &
\includegraphics[width=0.30\textwidth]{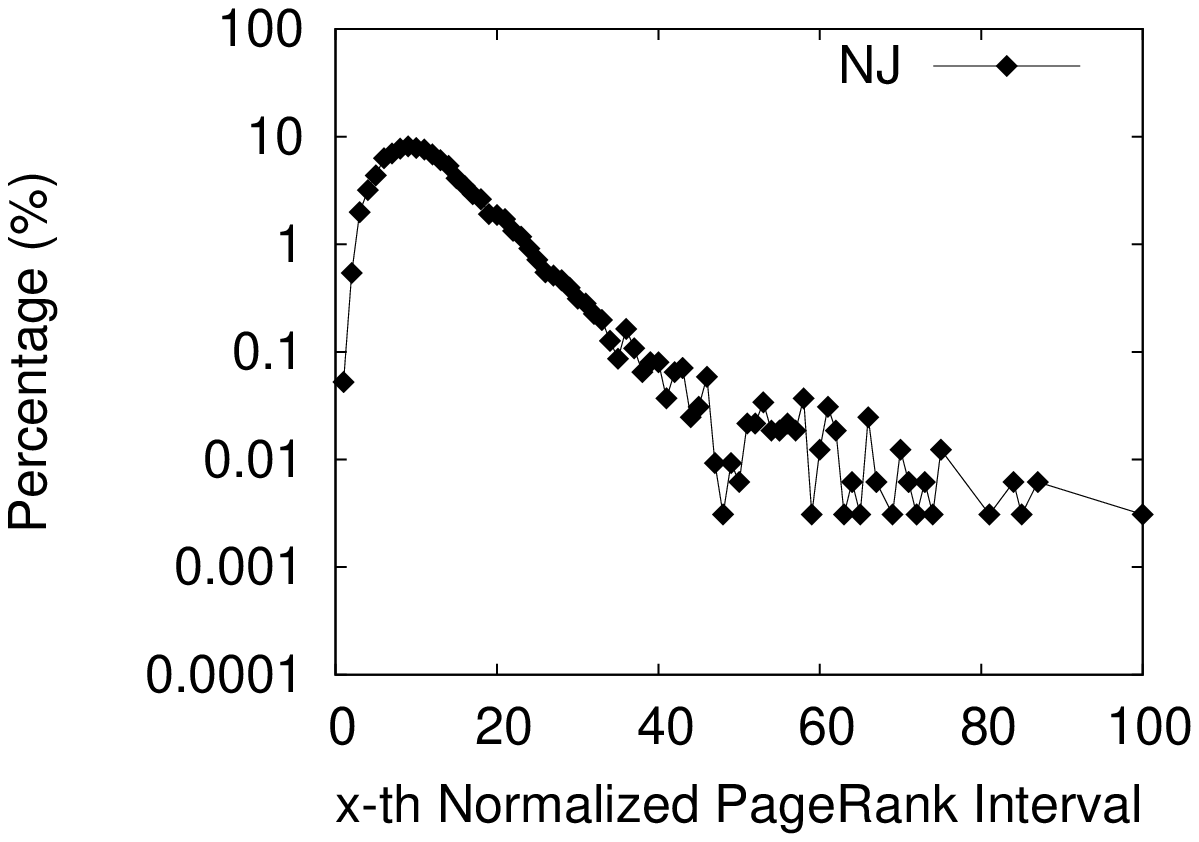}
    \\
    (a) \emph{WEB}
    &
    (b) \emph{CIT}
    &
    (c) \emph{NJ}
\end{tabular}
\caption{Distribution of Normalized PageRank Values}
\label{fig:PR}
\end{figure*}

Next, we divide the normalized PageRank values into 100 buckets, where
the $i$-th bucket indicates the interval $(i-1, i]$.
For example, the $10$-th bucket represents the interval $(9, 10]$.
We plot the buckets (on the x-axis) with respect to the percentage of
vertices whose normalized PageRank values are in the buckets (on the
y-axis) for the three graphs in Figure~\ref{fig:PR}.

As expected, the distribution of normalized Page\-Rank values in
\emph{NJ} is substantially different from those in \emph{WEB} and
\emph{CIT}.
Most of the vertices in \emph{WEB} and \emph{CIT} have very small
normalized PageRank values, and a small fraction of the vertices
spread over the remaining buckets, which is consistent with the degree
distributions shown in Figures~\ref{fig:indegree}
and~\ref{fig:Outdegree}.
For example, 98.8\% of the vertices in \emph{WEB} have normalized
PageRank values falling into the first bucket, i.e., $(0, 1]$; and
99.9\% of the vertices in \emph{WEB} have normalized PageRank values
falling into the first 5 buckets.

Such a distribution is good for ranking web pages because users only
care about a small fraction of the most important web pages. For
example, the PageRank values of the remaining 0.1\% of web pages
(which have larger PageRank values and thus are considered as
important) are spread over the remaining 95 buckets, making the
PageRank based ranking very discriminative for these web pages.

The normalized PageRank values in \emph{NJ} are distributed more
uniformly when compared with \emph{WEB} and \emph{CIT}. This
characteristic is bad for ranking: more vertices have the same or very
similar PageRank values, rendering the values much less discriminating
when compared to \emph{WEB} and \emph{CIT}.
Although this distribution reduces the usability of pageRank values
for ranking, the distribution does not reduce the utility of the
PageRank values for identifying similar road segments. In the paper,
we do not use PageRank values for ranking, but rather use PageRank
values for identifying road segments that may have similar traffic
flows.

Recall that if a road segment is covered by a training data set, our
approach (in particular, with the help of objective function $F_2$) is
able to propagate the obtained travel cost information to all the road
segments with similar PageRank values.
The kind of distribution observed on \emph{NJ} yields benefits in our
setting. For example, if a road segment is covered by a training data
set, the corresponding bucket (based on its PageRank value) may
contain relatively more segments, and thus the obtained travel cost
can be propagated to more segments. In contrast, with the
distributions observed in \emph{WEB} and \emph{CIT}, unless the
segment falls into the first few buckets, the obtained travel cost can
only be propagated to few segments.

\noindent
\textbf{Additional Related Work About Using PageRank on Road
  Networks:} Studies exist of the correlations between human movement
flows and various geometrical and topological
metrics~\cite{DBLP:journals/gis/Jiang09, jiang2008self}.
One study \cite{DBLP:journals/gis/Jiang09} considers human movement
(both pedestrians and vehicles) in London and three central regions in
London; another~\cite{jiang2008self} considers vehicle movement (based
on Annual Average Daily Traffic obtained from the Swedish Road
Administration and GPS tracks) in seven regions in Sweden.
Both studies suggest that weighted PageRank values have the best
correlation with traffic flows.
Further, both random and purposeful movements (movements for given
source-destination pairs) are simulated~\cite{jiang2011agent}, and
both movements generate almost the same aggregate traffic flows, which
are highly correlated with PageRank values. A live demo for the
simulation is
available\footnote{\url{http://fromto.hig.se/~bjg/movingbehavior/}}.

Another study~\cite{crisostomi2011google} uses the PageRank idea to
model traffic on a road network as a Markov chain.  The effectiveness
of the model is validated by a well-known traffic simulator,
SUMO~\cite{krajzewicz2002sumo, krajzewicz2006open}.
PageRank values are also used to predict the congestion levels in a
road network~\cite{pop2012efficient}, which is applied to optimize the
control of traffic lights.

A benchmark~\cite{DBLP:conf/petra/GeorgatzisP12} is developed to study
the Page\-Rank values on graphs representing the web, citations, and
road networks with emphasis on convergence speed.  The best
convergence speed is achieved on the graph representing a road
network.

Further, PageRank is used for ranking groups of places in a road
network~\cite{agryzkov2012algorithm}. The results show that,
typically, a group of places have the same or highly similar PageRank
values, which is consistent with our study on the \emph{NJ} dataset.

As a final remark, most of the works using Page\-Rank on road networks
are published in the recent three years, which indicates that this is
a relatively new research direction that is not yet well studied. No
existing works employ PageRank for predicting travel costs---our study
is the first to explore this.

\subsection{Adjacency Constraint}
\label{ssec:objadj}

The $\mathit{PRTC}$ function is derived from the overall structure of
the road network. In this section, we consider a finer-grained
topological aspect of the road network, namely, \textbf{directional
  adjacency}.

An important feature of a road network is that an event at one road
segment may propagate to influence adjacent road
segments. 
Consider a typical event in a road network, e.g., traffic congestion.
If congestion occurs on road
segment $(A, B)$ in Fig.~\ref{fig:roadnetwork}, road segment
$(B, C)$ may also experience congestion, or at least the traffic on $(B,
C)$ is affected by the congestion that occurs on $(A, B)$.
Thus, the cost variables of two directionally adjacent road segments
should be similar.

The directional adjacency we discus here is represented clearly in the
dual graph. If and only if two dual vertices are connected by an dual
edge in the dual graph, the two corresponding road segments are
directionally adjacent.
For example, although edges $(B$, $D)$ and $(B$, $C)$ (in
Fig.~\ref{fig:primal}) intersect, their cost variables may not
necessarily tend to be similar because no vehicle can travel between
these two edges. Directional adjacency is distinct from the
``non-directional'' adjacency considered in previous
work~\cite{DBLP:conf/aaai/IdeS11}.

Another point worth noting is that if two road segments represent
opposite directions of the same physical road segment, they are not
directionally adjacent.
It is natural that an event on a physical road only yields congestion
in one direction, but not both directions.
Considering the edges $(A, B)$ and $(B, A)$ (in
Fig.~\ref{fig:primal}), their corresponding vertices in the dual
graph ($AB$ and $BA$ in Fig.~\ref{fig:dual}) are connected by two
edges, however, their cost variables are not necessarily similar.

Directional adjacency is also temporally sensitive. For example,
although edges $(A, B)$ and $(B, C)$ are directionally adjacent,
the general traffic situation (indicated by the cost variable) on
edge $(A, B)$ during peak hours is not necessarily correlated
with the traffic on edge $(B, C)$ during non-peak hours.

To incorporate directional adjacency, we incorporate a Directionally
Adjacent Temporal Constraint ($\mathit{DATC}$) function into the
overall objective function.

\begin{equation}
\small
\label{eq:DATC} \mathit{DATC} (\mathbf{d}) = \sum_{k=1}^{k=|\mathit{TAGS}|}
\mathit{DATC}(\mathbf{d}, k),
\end{equation}
where
\begin{displaymath}
\begin{small}
\mathit{DATC}(\mathbf{d}, k) = \sum_{i,j=1}^{|G.\mathbb{E}|}
W_k'(v_{e_i}', v_{e_j}') \cdot (d_{(e_i, tag_k)}-d_{(e_j,
tag_k)})^2,
\end{small}
\end{displaymath}
and where $v_{e_i}'$ and $v_{e_j}'$ have the same meaning as in
Equation~\ref{eq:topsim}. $W_k'(v_{e_i}', v_{e_j}')$ is as defined
in Equation~\ref{eq:weightTime} if $v_{e_i}'$ and $v_{e_j}'$ do not
indicate the same physical road segment; and $W_k'(v_{e_i}',
v_{e_j}')$ equals 0 otherwise.
For instance, although $W_{\mathit{PEAK}}(AB, BA)=\frac{1}{43}$ as
discussed in Section~\ref{ssec:wpc}, $W_{\mathit{PEAK}}'(AB, BA)=0$
since $AB$ and $BA$ indicate the same physical road segment,
\emph{Avenue}~1.

The $\mathit{DATC}$ function aims to make the cost variables satisfy
the following property: given road segments $e_i$ and $e_j$, if a many
of the trips that follow $e_i$ also follow $e_j$, as indicated by
$W_k'(v_{e_i}', v_{e_j}')$, the cost variables on the two edges tend
to be more correlated.

Similar to the discussion in Section~\ref{sssec:PRTC}, we introduce a
block diagonal matrix $\mathbf{B}\in
\mathbb{\mathbb{R}}^{|\mathbf{d}|\times |\mathbf{d}|}$ with the same
format as matrix $\mathbf{A}$ (defined in Equation~\ref{eq:A}).
In particular, in each block matrix, $\mathbf{B_k}[i, j]=\max(
W_k'(v_{e_i}', v_{e_j}')$, $W_k'(v_{e_i}', v_{e_j}'))$, which
guarantees that matrix $\mathbf{B_k}$, and hence matrix $\mathbf{B}$,
are symmetric.
Note that it is not possible that both $W_k'(v_{e_i}', v_{e_j}')$ and
$W_k'(v_{e_j}', v_{e_i}')$ are non-zero because if edge
$\mathit{D2P}(v_{e_i}')$ is directionally adjacent to edge
$\mathit{D2P}(v_{e_j}')$ then edge $\mathit{D2P}(v_{e_j}')$ cannot be
directionally adjacent to edge $\mathit{D2P}(v_{e_i}')$.
Let $\mathbf{L_B}$ to be the graph Laplacian derived by matrix
$\mathbf{B}$.
The $\mathit{DATC}$ function is represented by
Equation~\ref{eq:DATCMatrix}.
\begin{equation}
\small \label{eq:DATCMatrix} \mathit{DATC}(\mathbf{d}) =
\mathbf{d}^\mathbf{T}\mathbf{L_B} \mathbf{d}
\end{equation}

\subsection{Solving The Problem}

Combining the three individual objective functions and a classical
$\mathit{L2}$ regularizer, we obtain the overall objective function
$O(\mathbf{d})$:
\[
\begin{small}
\label{eq:overall} O(\mathbf{d}) = \mathit{RSS}(\mathbf{d}) + \alpha \cdot
\mathit{PRTC}(\mathbf{d}) + \beta \cdot \mathit{DATC}(\mathbf{d})+ \gamma \cdot
||\mathbf{d}||_2^2,
\end{small}
\]
where $\alpha$, $\beta$, and $\gamma$ are hyper-parameters that
control the tradeoff among the losses on $\mathit{RSS}$,
$\mathit{PRTC}$, $\mathit{DATC}$, and the $\mathit{L2}$ regularizer.
The matrix representation of the objective function is shown in
Equation~\ref{eq:overallMatrix}.
\begin{equation}
\small \label{eq:overallMatrix} O(\mathbf{d}) =
||\mathbf{c} - \mathbf{Q}^\mathbf{T} \mathbf{d}  ||_2^2 +\alpha \cdot
\mathbf{d}^\mathbf{T} \mathbf{L_A} \mathbf{d} +\beta \cdot
\mathbf{d}^\mathbf{T} \mathbf{L_B} \mathbf{d} + \gamma \cdot
||\mathbf{d}||_2^2
\end{equation}
By differentiating Equation~\ref{eq:overallMatrix} w.r.t.\
vector $\mathbf{d}$ and setting it to $0$, we get
\begin{equation}
\small
\label{eq:diff} [\mathbf{Q}\mathbf{Q}^\mathbf{T}+\alpha \cdot \mathbf{L_A}
+ \beta \cdot \mathbf{L_B} + \gamma \cdot \mathbf{I}] \mathbf{d}
=\mathbf{Q}\mathbf{c}.
\end{equation}
The solution to Equation~\ref{eq:diff} is the optimal solution to the
cost vector, denoted as $\mathbf{\widehat{d}}$, that minimizes the
overall objective function in Equation~\ref{eq:overallMatrix}.
The linear system in Equation~\ref{eq:diff} can be solved efficiently
by several iterative algorithms such as the conjugate gradient
algorithm~\cite{golub1996matrix}.

Finally, feeding the optimized cost variable vector
$\mathbf{\widehat{d}}$ to function $G.H$, the time varying weights of
the graph become available.

\subsection{Discussion}

In addition to the topology of a road network, other aspects of edges
may be useful for identifying similarities among edges, e.g., the
shapes and capacities of edges and the points of interest along
edges~\cite{yuan2012t}.
Such information is not always available in digital maps and can be
difficult to obtain. However, it is of interest to extend the proposed
methods to take additional information, when available, into account.
To achieve general applicability of the paper's methods, we minimize
the requirements of the input graph $G''$: both $\mathit{PRTC}$ and
$\mathit{DATC}$ rely solely on the topology of a road network, which
can be obtained easily from any digital map.

The weight annotation problem is finally handled by solving a
system of linear equations, i.e., Equation~\ref{eq:diff}.
Alternative edge similarity metrics (e.g., considering the shapes and
capacities of edges) can be easily incorporated into the linear system
by adding new terms of the form $\varphi\cdot \mathbf{L_M}$, where
$\varphi$ is the hyper-parameter and $\mathbf{L_M}$ is the Laplacian
matrix derived by an alternative similarity metric.
An alternative similarity metric $\mathit{sim}$ should satisfy
symmetry: $\mathit{sim}(e_i, e_j)=\mathit{sim}(e_j, e_i)$. Both
$\mathit{PRTC}$ and $\mathit{DATC}$ satisfy symmetry.

The core operations in solving a system of linear equations using a
conjugate gradient algorithm are matrix multiplication and
transposition.
This means that existing scalable matrix computation
algorithms~\cite{seo2010hama,lin2010data} can be applied directly to
make the proposed framework scalable and applicable to large road
networks.

\section{Experimental Study}
\label{sec:exp}

We study the effectiveness of the proposed method for weight
annotation of road networks with both \emph{travel time} (\emph{TTWA})
and \emph{GHG emissions} (\emph{GEWA}).

\subsection{Experimental Setup}

\textbf{Road Networks:} We use two road networks. The SK network is
from Skagen, Denmark and has a primal graph with $543$ vertices and
$1,244$ edges.
The NJ network contains almost all of North Jutland, Denmark and has a
primal graph with $17,956$ vertices and $39,372$ edges.

\textbf{Trips:} We use GPS observations collected from $28$ vehicles
in the period 2007-10-01 to 2007-10-15. When the vehicles were moving,
positions were sampled at 1~Hz.
The data is collected as part of an experiment where young drivers
start out with a substantial rebate on their car insurance and then
are warned if they exceed the speed limit and are penalized
financially if they continue to speed.

We apply an existing tool for map matching GPS observations onto road
segments, thus obtaining $431$ trips in the SK network and $11,516$
trips in the NJ network.

For \emph{TTWA}, we use the total travel time for each trip, which can
be obtained directly from the GPS observations of the trip, as the
cost.

For \emph{GEWA}, we use the GHG emissions of each trip as trip
cost. Ideally, the exact fuel consumption should be obtained from CAN
bus sensor data. Since such data is hard to obtain in a scalable
fashion, we use instead the VT-micro model~\cite{ahn2002estimating}
that is able to compute the GHG emissions of trips based on the
instantaneous velocities and accelerations derived from the GPS
records of the trips in a robust fashion~\cite{guoecomark}. The 1~Hz
GPS sampling frequency makes the VT-Micro model easy to use.

\textbf{Traffic Category Tags:}
In transporation research, \emph{PEAK} and \emph{OFFPEAK} periods are
used widely to distinguish different traffic flows over the course of
a day~\cite{cantos2011viability}.
Thus, we use \emph{PEAK} and \emph{OFFPEAK} as traffic category tags.
Further, we distinguish between weekdays from weekend days, as traffic
differs between weekdays and weekend days.
To appropriately assign \emph{PEAK} and \emph{OFFPEAK} tags to the
data set, we plot the numbers of GPS records according to their
corresponding observed time at an one-hour granularity for weekdays
and weekend days, respectively.
Based on the generated histograms, we identify \emph{PEAK} and
\emph{OFFPEAK} periods for weekdays. We find no clear peak periods
during weekends and thus use \emph{WEEKENDS} as the single tag for
weekends.
Table~\ref{table:temporaltags} provides the mapping (i.e., the
function $G.F$) from time periods to tags.

\begin{table}[!htbp]
\caption{Traffic Category Tag Function $G.F$}
\label{table:temporaltags}
\small \centering
\begin{tabular}{lll}
\hline\noalign{\smallskip}
 Periods &  &  Tags   \\
\noalign{\smallskip}\hline\noalign{\smallskip}
Weekdays & [0:00, 7:00) & \emph{OFFPEAK}   \\
Weekdays & [7:00, 8:00) & \emph{PEAK}      \\
Weekdays & [8:00, 15:00) & \emph{OFFPEAK}  \\
Weekdays & [15:00, 17:00) & \emph{PEAK}    \\
Weekdays & [17:00, 24:00) & \emph{OFFPEAK} \\
Weekends & [0:00, 24:00) & \emph{WEEKENDS} \\
\noalign{\smallskip}\hline
\end{tabular}
\end{table}

T-Drive~\cite{DBLP:conf/gis/YuanZZXXSH10} is able to assign distinct
and fine-grained traffic tags to individual edges. The precondition of
the method is that sufficient GPS data is associated with
edges. However, a substantial fraction of all edges have no GPS data
in our setting. Thus, we use traffic tags at the coarse granularity
shown in Table~\ref{table:temporaltags}.

\textbf{Implementation Details:} The PageRank computation is
implemented in C using the iGraph library
version~0.5.4~\cite{iGraph}. 
All remaining experiments are implemented in Java, where the conjugate
gradient algorithm for solving a linear system is implemented using
the MTJ (matrix-toolkits-java)
package~\cite{matrixjava}. 

We use the threshold 0.95 to filter the entries in the PageRank-based
similarity matrix $\mathbf{A}$ (Equation~\ref{eq:A}): if the value of
an entry in $\mathbf{A}$ is smaller than 0.95, the entry is set to 0.
We use the speed limits associated with roads to classify the edges
into two categories, \emph{highways} (with speed limits above $90$
km/h) and \emph{urban roads} (with speed limits below $90$ km/h). We
only apply adjacency constraint on pairs of edges in the same
category.

Due to the space limitation, the experiments only report the results
using the best set of hyper-parameters, which are is obtained by
manual tuning on a separate data set using cross validation. This is a
well known method~\cite{bishop2006pattern} for choosing
hyper-parameters.

\subsection{Experimental Results}

\subsubsection{Effectiveness Measurements}

To gain insight into the accuracy of the obtained trip cost based
weights, we split the set of $(\mathit{trip}$, $\mathit{cost})$ pairs
into a training set $\mathbb{TC}_{\mathit{train}}$ and a testing set
$\mathbb{TC}_{\mathit{test}}$.
We use the the training set to annotate the spatial network with
weights, and we use the the testing set to evaluate the accuracy of
the weights.
In the following experiments, we randomly choose 50\% of the pairs for
training and the remaining 50\% for testing, unless explicitly stated
otherwise.

Since no ground-truth time-dependent weights exist for the two road
networks, the accuracy of the obtained weights can only be evaluated
using the trips in testing set $\mathbb{TC}_{test}$.
If the obtained weights (using $\mathbb{TC}_{train}$) actually reflect
the travel costs, the difference between the actual cost and the
estimated cost using the obtained weights (i.e., by using
Equation~\ref{eq:cost_trip} defined in Section~\ref{ssec:trip}) for
each trip in the testing set $\mathbb{TC}_{test}$ should be small.

We use the sum of squared loss ($\mathit{SSL}$) value (defined in
Equation~\ref{eq:ssl}) between the actual cost $c^{(i)}$ and the
estimated cost $cost(t^{(i)})$ over every trip in the testing set
$\mathbb{TC}_{\mathit{test}}$ to measure the accuracy of the obtained
weights.
\begin{equation}
\small \label{eq:ssl}
\mathit{SSL}(\mathbb{TC}_{\mathit{test}})=\sum_{(t^{(i)}, c^{(i)})
\in \mathbb{TC}_{\mathit{test}}} (c^{(i)} - cost(t^{(i)}))^2
\end{equation}
For example, if the GHG emissions based weights really reflect the
actual GHG emissions, the sum of squared loss between the actual GHG
emissions and the estimated GHG emissions over every testing trip
should tend to be small. The smaller the sum of squared loss, the
more accurate the weights.

To gain insight into the effectiveness of the proposed objective
functions, we compare four combinations of the functions:
\begin{enumerate}
  \item $F_1$=$\mathit{RSS}(\mathbf{d})$ + $\gamma \cdot ||\mathbf{d}||_2^2$.
  \item $F_2$=$\mathit{RSS}(\mathbf{d}) + \alpha \cdot
\mathit{PRTC}(\mathbf{d})$ + $\gamma \cdot ||\mathbf{d}||_2^2$.
    \item $F_3$=$\mathit{RSS}(\mathbf{d}) + \beta \cdot
\mathit{DATC}(\mathbf{d})$ + $\gamma \cdot ||\mathbf{d}||_2^2$.
  \item $F_4$=$\mathit{RSS}(\mathbf{d}) + \alpha \cdot
\mathit{PRTC}(\mathbf{d}) + \beta \cdot \mathit{DATC}(\mathbf{d})$+
$\gamma \cdot ||\mathbf{d}||_2^2$.
\end{enumerate}
Function $F_1$ only considers the residual sum of squares. Functions
$F_2$ and $F_3$ take into account the PageRank-based topological
constraint and the directional adjacency constraint,
respectively. Function $F_4$ takes into account both constraints.

As the objective function used in trajectory
regression~\cite{DBLP:conf/aaai/IdeS11} also considers adjacency, we
can view the method using function $F_3$ as an improved version of
trajectory regression because (\rmnum{1}) function $F_3$ works not
only for travel times, but also other travel costs, e.g., GHG emissions;
(\rmnum{2}) function $F_3$ considers the temporal variations of travel
costs, while trajectory regression does not; and (\rmnum{3}) function
$F_3$ considers directional adjacency, while trajectory regression
models a road network as a undirected graph and only considers
undirected adjacency.

The sum of squared loss value for using objective function $F_i$ is
denoted as $\mathit{SSL}_{F_i}(\mathbb{TC}_{\mathit{test}})$.
In order to show the relative effectiveness of the proposed
objective functions, we report the ratios $\mathit{Ratio}_{F_2}$=$\frac{\mathit{SSL}_{F_2}(\mathbb{TC}_{\mathit{test}})}{\mathit{SSL}_{F_1}(\mathbb{TC}_{\mathit{test}})}$,
$\mathit{Ratio}_{F_3}$=$\frac{\mathit{SSL}_{F_3}(\mathbb{TC}_{\mathit{test}})}{\mathit{SSL}_{F_1}(\mathbb{TC}_{\mathit{test}})}$,
and
$\mathit{Ratio}_{F_4}$=$\frac{\mathit{SSL}_{F_4}(\mathbb{TC}_{\mathit{test}})}{\mathit{SSL}_{F_1}(\mathbb{TC}_{\mathit{test}})}$.

Coverage, defined in Equation~\ref{eq:coverage}, is introduced as
another measurement.
\begin{equation}
\small  \label{eq:coverage}
\mathit{Cove}_{F_i}(\mathbb{TC}_{\mathit{train}})=\frac{|\{ e|e\in
G.\mathbb{E} \wedge \mathit{annotated}(e) \}|}{|G.\mathbb{E}|},
\end{equation}
where $\mathit{annotated}(e)$ holds if edge $e$ is annotated with
weights using
$\mathbb{TC}_{\mathit{train}}$.
Function $\mathit{Cove}_{F_i}$ indicates the ratio of the number of
edges whose weights have been annotated by using objective function
$F_i$ to the total number of edges in the road network.
The higher the coverage is, the more edges in the road network are
annotated with weights, and thus the better performance.

\subsubsection{Travel Time Based Weight Annotation}

\textbf{Effectiveness of objective functions:}
Table~\ref{tbl:timeratio} reports the results on travel time
based weight annotation.
Column $\mathit{SSL}_{F_1}$ reports the absolute $\mathit{SSL}$ values
over all test trips when using objective function $F_1$ for both data
sets.
NJ has much larger $\mathit{SSL}$ values than SK because it has much
more testing trips.
For both road networks, the weights annotated using objective
function $F_4$ have the least $\mathit{SSL}$ values.

\begin{table}[hbt]
\caption{ Effectiveness on \emph{TTWA}}
\label{tbl:timeratio}
\centering
\small
\begin{tabular}{|l|l|l|l|l|} \hline
 & $\mathit{SSL}_{F_1}$ & $Ratio_{F_2}$ & $Ratio_{F_3}$ & $Ratio_{F_4}$ \\ \hline
SK & 88,656 & 99.2\% & 44.0\% & 43.8\% \\ \hline
NJ & 14,823,752 & 92.2\% & 49.2\% & 43.1\% \\ \hline
\end{tabular}
\end{table}

We also observe that the PageRank based topological constraint works
more effectively on NJ than on SK.
The reason is that Skagen is a small town in which few road segments
have similar topology (e.g., similar weighted PageRank values).
In the NJ network, the PageRank based topological constraint gives a
better accuracy improvement since more road segments have similarly
weighted PageRank values.

The coverage reported in Table~\ref{tbl:coverage} also justifies the
observation.
When using objective function $F_1$, only the edges in the set of
training trips can be annotated, which can be expected to be a small
portion of the road network.
\begin{table}[hbt]
\caption{ Coverage of Weight Annotation} \label{tbl:coverage}
\centering
\small
\begin{tabular}{|l|l|l|l|l|} \hline
 & $Cove_{F_1}$ & $Cove_{F_2}$ & $Cove_{F_3}$ & $Cove_{F_4}$\\ \hline
SK  & 22.8\% & 28.8\% & 100\% & 100\%\\ \hline
NJ &34.8\% & 86.7\% & 99.6\% & 100\%\\ \hline
\end{tabular}
\end{table}
When using objective function $F_2$, the coverage of the SK network
increases much less than for the NJ network.
This suggests that in a large road network, the PageRank based
topological constraint substantially increases the coverage of the
annotation, thus improving the overall annotation accuracy.

The directed adjacency topological constraint yields similar accuracy
improvements on both road networks, and the accuracy improvement is
more substantial than the improvement given by the PageRank based
topological constraint.
This is as expected because a road network is fully connected, and
$\mathit{DATC}$ is able to finally affect almost every edge, which
gives more information for the edges that are not traversed by trips
in the training set. This can be observed from the third column of
Table~\ref{tbl:coverage}.

For both road networks, $\mathit{PRTC}$ and $\mathit{DATC}$ together
give the best accuracy, as shown in column $\mathit{Ratio}_{F_4}$ in
Table~\ref{tbl:timeratio}. This finding offers evidence of the
overall effectiveness of the proposed objective functions.

\textbf{Accuracy comparison with a baseline:}
The test tips contain edges that are not covered by any training
trips. Therefore, existing methods~\cite{DBLP:conf/gis/YuanZZXXSH10}
that can estimate travel time based on historical data are
inapplicable as baseline.

If the speed limit of every edge in a road network is available, we
can use speed limit derived weights as a baseline for travel time
based weight annotation.
While it is difficult to obtain a speed limit for every road segment
in a road network, we can use default values were values are
missing. In the NJ network, $62$ edges lack a speed limit and are
assigned a default value (50~km/h).


Given an edge $e$ and its speed limit $sl(e)$ and length $G.L(e)$, the
corresponding travel time based weight for $e$ is $ \lambda \cdot
\frac{G.L(e)}{sl(e)}$ if $e$ is an urban road (where $\lambda \geq 1$)
and $\frac{G.L(e)}{sl(e)}$ if $e$ is a highway.

The factor $\lambda$ is used because vehicles tend to travel at speeds
below the speed limit on urban roads and at the speed limit on
highways.
Previous work~\cite{DBLP:conf/aaai/IdeS11} uses $\lambda=2$, meaning
that vehicles normally travel at half the speed limit in urban
regions. However, we find that $\lambda = 1$ works the best for our
data.
The reason may be two-fold: (\rmnum{1}) the data we use is collected
from young drivers who tend to drive more aggressively than average
drivers. (\rmnum{2}) the SK and NJ networks are relatively
congestion-free when compared to Kyoto, Japan, which is simulated in
previous work~\cite{DBLP:conf/aaai/IdeS11}.

The above allows us to treat the speed limit derived weights as a
baseline method for travel time based weight annotation.
To observe the accuracy of the baseline method, its accuracy is also
evaluated using $\mathit{SSL}$ over every testing trip. Specifically,
the baseline with $\lambda = 2$ is denoted as
$\mathit{SSL}_{BL,\lambda=2}(\mathbb{TC}_{\mathit{test}})$, and the
baseline with $\lambda = 1$ is denoted as
$\mathit{SSL}_{BL,\lambda=1}(\mathbb{TC}_{\mathit{test}})$.
The two resulting baselines are compared with the proposed method, and
the results are reported in Table~\ref{tbl:timebaseline}, where
$\mathit{Ratio}_{\lambda=2}$=$\frac{\mathit{SSL}_{F_4}(\mathbb{TC}_{\mathit{test}})}{\mathit{SSL}_{BL,\lambda=2}(\mathbb{TC}_{\mathit{test}})}$
and
$\mathit{Ratio}_{\lambda=1}$=$\frac{\mathit{SSL}_{F_4}(\mathbb{TC}_{\mathit{test}})}{\mathit{SSL}_{BL,\lambda=1}(\mathbb{TC}_{\mathit{test}})}$.
The ratios $\mathit{Ratio}_{\lambda=1}$ on the two road networks show
that the weights obtained by our method are substantially better than
the best cases of the weight obtained from the speed limits.

\begin{table}[hbt]
\caption{Comparison With Baselines on \emph{TTWA}}
\label{tbl:timebaseline}
\centering
\small
\begin{tabular}{|l|l|l|} \hline
 & $\mathit{Ratio}_{\lambda=2}$ & $\mathit{Ratio}_{\lambda=1}$  \\ \hline
SK  & 36.0\% & 78.8\%  \\ \hline
NJ  & 24.2\% & 90.8\% \\ \hline
\end{tabular}
\end{table}

\begin{figure}[htbp]
\centering
\begin{tabular}{@{}c@{}c@{}c@{}}
    \includegraphics[width=0.5\columnwidth]{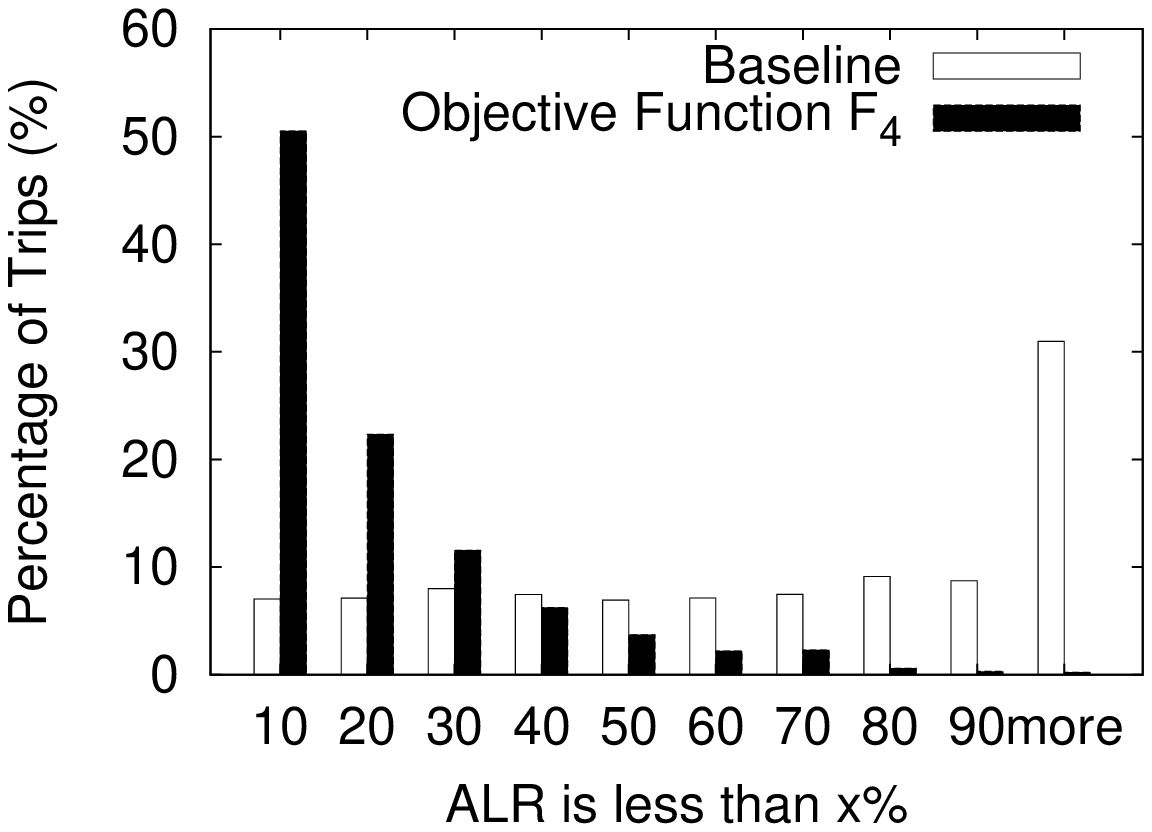}
    &
    \includegraphics[width=0.5\columnwidth]{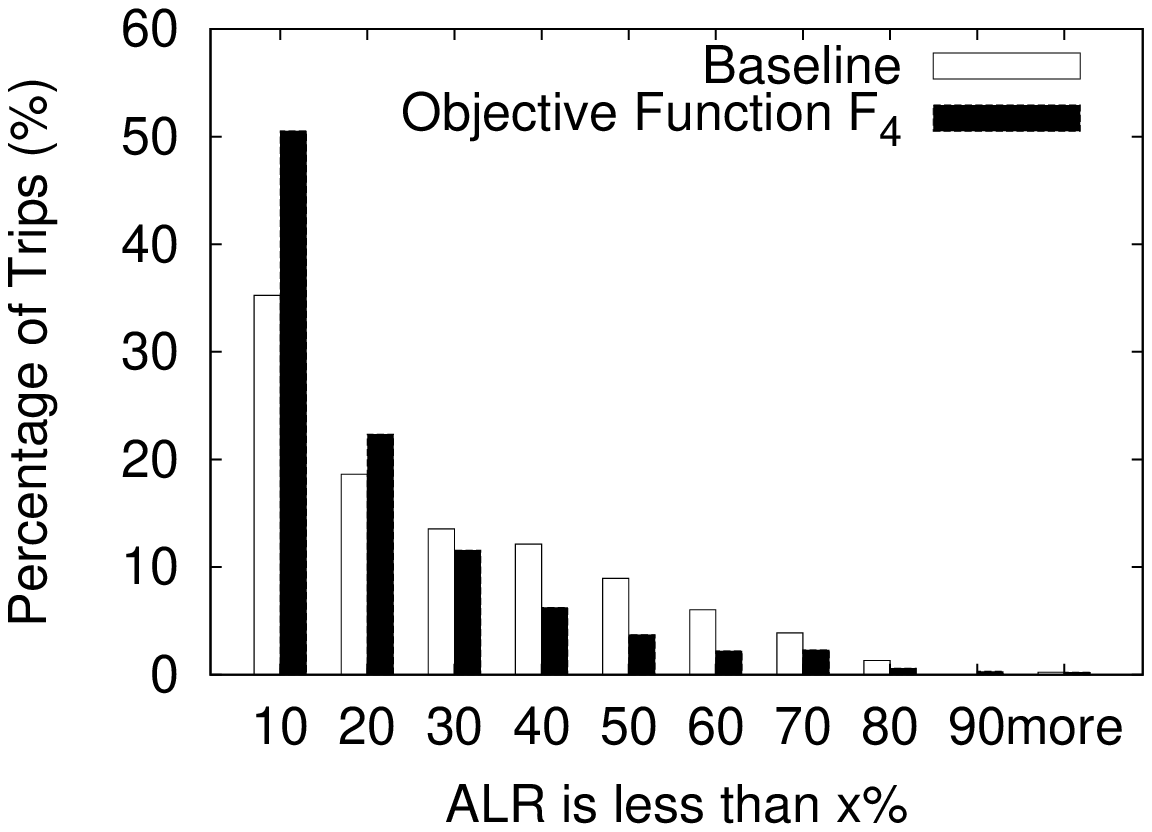}
    \\
    \small(a) Baseline with $\lambda=2$
    &
    \small(b) Baseline with $\lambda=1$
\end{tabular}
\caption{$\mathit{ALR}$ Comparison on \emph{TTWA} of NJ} \label{fig:compTime}
\end{figure}

The same deviation has quite a different meaning for long versus short
trips. For example, a 50-second deviation can be considered as a very
good estimation error for a 30-minute trip, while it is a poor
estimation error for a 2-minute trip.
Thus, to better understand how the overall $\mathit{SSL}$ values are
distributed, we plot the number of test trips whose \emph{absolute
  loss ratio} ($\mathit{ALR}$) values are within $x$ percentage in
Fig.~\ref{fig:compTime}.
Given a test pair $(t^{(i)}$, $c^{(i)}) \in
\mathbb{TC}_{\mathit{test}}$, its $\mathit{ALR}$ value equals the
absolute difference between the estimated and actual costs divided by
the actual cost, as defined in Equation~\ref{eq:ALR}.
\begin{equation}
\small \label{eq:ALR}
\mathit{ALR}((t^{(i)}, c^{(i)})) = \frac{\mathit{absolute}(cost(t^{(i)})-c^{(i)})}{c^{(i)}}
\end{equation}
Our method shows the best result as the majority of the test trips
have smaller $\mathit{ALR}$ values.
Assume that we consider and $\mathit{ALR}$ below 30\% as a good
estimation. Fig.~\ref{fig:compTime} shows that 84.3\% of test trips
have good estimations using the proposed method. In contrast, only
67.4\% and 22.1\% of test trips have good estimations using baseline
methods with $\lambda=1$ and $\lambda=2$, respectively.

We do not integrate speed limits into our method because (\rmnum{1})
for edges without available speed limits, the obtained weights are
quite sensitive to the assigned default speed limits: inaccurate
defaults deteriorate the performance severely; and (\rmnum{2}) speed
limits do not give obvious benefits when annotating edges with GHG
emissions based weights, as we will see shortly in
Section~\ref{sssec:GHGexp} (in particular, in Fig.~\ref{fig:compFuel}).

\subsubsection{GHG Emissions Based Weight Annotation}
\label{sssec:GHGexp}

\textbf{Effectiveness of objective functions:}
Table~\ref{tbl:fuelratio} reports the results on GHG emissions
based weight annotation.
\begin{table}[hbt]
\caption{Effectiveness on \emph{GEWA}}
\label{tbl:fuelratio}
\centering
\small
\begin{tabular}{|l|l|l|l|l|} \hline
 & $\mathit{SSL}_{F_1}$ & $\mathit{Ratio}_{F_2}$ & $\mathit{Ratio}_{F_3}$ & $\mathit{Ratio}_{F_4}$\\ \hline
SK & 175.931 & 99.9\% & 40.3\% & 30.0\% \\ \hline
NJ & 87,362,465 & 94.5\%& 66.2\% & 44.3\% \\ \hline
\end{tabular}
\end{table}
In general, the results are consistent with the results from the
travel time based weight annotation (as shown in
Table~\ref{tbl:timeratio}):
(\rmnum{1}) The PageRank-based topological constraint works more
effectively on the NJ network than on the SK network;
(\rmnum{2}) the directed adjacency constraint works more
effectively than the PageRank-based topological constraint;
(\rmnum{3}) the weights obtained by using both $\mathit{PRTC}$ and
$\mathit{DATC}$ give the best accuracy.
The coverage when using the different objective functions is exactly
the same as what was reported in Table~\ref{tbl:coverage}.

\textbf{Comparison with a baseline:} As we did for travel times, we
use speed limits to devise a baseline for GHG emissions based weight
annotation.
Assuming a vehicle travels on an edge at constant speed (e.g., the
speed limit of the edge), we can simulate a sequence of instantaneous
velocities.
For example, let an edge be 100~meters long and the speed limit be
60~km/h.  The simulated trip on the road segment is represented by a
sequence of 6 records, each with 60~km/h as the instantaneous
velocity.
This allows us to apply the VT-micro model to estimate GHG emissions
based edge weights.
Since in the previous set of experiments, we have already found that
the speed limit (i.e., $\lambda=1$) is the best fit for our data we
simply use the speed limit here.

\begin{figure}[h]
\centering
  \includegraphics[width=0.6\columnwidth]{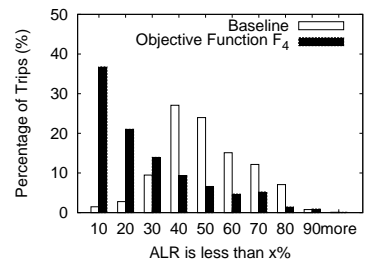}
\caption{ $\mathit{ALR}$ Comparison on \emph{GEWA} of NJ} \label{fig:compFuel}
\end{figure}

We obtain $\mathit{Ratio}_{\lambda=1} = 24.7\%$ for SK and
$\mathit{Ratio}_{\lambda=1} = 29.8\%$ on NJ.
Fig.~\ref{fig:compFuel} shows the percentage of test trips whose
$\mathit{ALR}$ values are less than $x\%$ using the baseline with
$\lambda=1$ and the proposed method, respectively.
These results clearly show the better performance of the proposed
method, as the majority of test trips have smaller $\mathit{ALR}$
values.

\subsubsection{Effectiveness of the Size of Training Trips}

\begin{figure}[ht]
\centering
  \includegraphics[width=0.6\columnwidth]{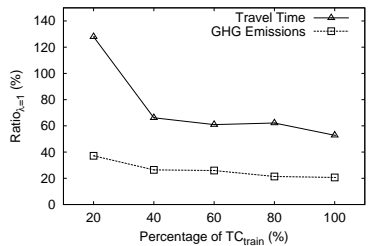}
\caption{Results on Different Size of $\mathbb{TC}_{train}$}
\label{fig:size}       
\end{figure}

In this section, we study the accuracy when varying the training set
size.
Specifically, on the NJ network, we reserve 20\% of the
$(\mathit{trip}$, $\mathit{cost})$ pairs as the testing set, denoted
as $\mathbb{TC}_{\mathit{test}}$, and the remaining 80\% as the
training set, denoted as $\mathbb{TC}_{train}$.
In order to observe the accuracy of weight annotation on different
sizes of $\mathbb{TC}_{\mathit{train}}$, we use 100\%, 80\%, 60\%,
40\% and 20\% of $\mathbb{TC}_{\mathit{train}}$ to annotate the
weights, respectively. The results are shown in Fig.~\ref{fig:size}.

For travel time, when only 20\% of $\mathbb{TC}_{\mathit{train}}$ is
used, the accuracy of our method is worse than the baseline method
with $\lambda=1$ because the baseline has a rough estimation for the
costs of all edges, while the 20\% of $\mathbb{TC}_{\mathit{train}}$
covers only 16.3\% of the edges in the road network.
Although our method propagates weights to edges that are not covered
by the training trips, the accuracy suffers when the initial coverage
of the training trips is low.
When 40\% of $\mathbb{TC}_{\mathit{train}}$ is used, the accuracy of
our method is much better than that of the baseline. In this case, the
training trips cover 23.3\% of all edges.
As the training set size increases, the accuracy of the travel time
weights also increases.
When we use all trips in $\mathbb{TC}_{\mathit{train}}$, the accuracy
of our method is almost twice that of the baseline.

For GHG emissions, we observe a similar trend: with more training
trips, the accuracy of the corresponding weights improves, and our
method always outperforms the baseline when annotating edges with GHG
emissions based weights.

This experiment justifies that (\rmnum{1}) our method works
effectively even when the coverage of the trips in the training set is
low;
(\rmnum{2}) if the coverage of the trips in the training set
increases, e.g., by providing more $(\mathit{trip}$, $\mathit{cost})$
pairs as training set, the accuracy of the obtained weights also
increases.

\section{Conclusion and Outlook}
\label{sec:con}

Reduction in GHG emissions from transportation calls for effective
eco-routing, and road network graphs where all edges are annotated
with accurate weights that capture environmental costs, e.g., fuel
usage or GHG emissions, are needed for eco-routing.
However, such weights are not always readily available for a road
network.
This paper proposes a general framework that takes as input a
collection of $(\mathit{trip}$, $\mathit{cost})$ pairs and assigns
trip cost based weights to a graph representing a road network, where
trip cost based weights may reflect GHG emissions, fuel consumption,
or travel time.
By using the framework, edge weights capturing environmental impact
can be computed for the whole road network, thus enabling eco-routing.
To the best of our knowledge, this is the first work that provides a
general framework for assigning trip cost based edge weights based on
a set of $(\mathit{trip}$, $\mathit{cost})$ pairs.

Two directions for future work are of particular interest.  It is of
interest to explore whether accuracy improvement is possible by using
distinct \emph{PEAK} and \emph{OFFPEAK} tags for different road
segments. Likewise, it is of interest to explore means of updating
weights in real time. A module that takes as input real time streaming
data, e.g., real time GPS observations along with costs, can be
incorporated into the framework.

\section*{Acknowledgments}
\label{sec:ack}

This work was supported by the Reduction project that is funded by the
European Commission as FP7-ICT-2011-7 STREP project number 288254.

\bibliographystyle{unsrt}
\end{document}